\documentclass[10pt]{iopart}

\usepackage{newtxmath}
\usepackage{colortbl,multirow,tabularx}
\usepackage{ragged2e}
\usepackage{subcaption}
\usepackage{multirow}
\usepackage{graphicx}
\usepackage{acronym}
\usepackage{xcolor}
\usepackage[abbr,dcu]{harvard}
\usepackage{lscape}
\usepackage{booktabs}
\usepackage{longtable}
\usepackage{comment}
\usepackage{float}
\usepackage{siunitx}

\usepackage{changes}

\graphicspath{{img/} {./}}

\definecolor{gray}{rgb}{0.25,0.5,0.5}

\usepackage{changes}

\acrodef{ANN}[ANN]{Artificial Neural Network}
\acrodef{BCM}[BCM]{Bienenstock Cooper Munro}
\acrodef{BP}[BP]{Back-Propagation}
\acrodef{BPTT}[BPTT]{Back-Propagation Through Time}
\acrodef{CMOS}[CMOS]{Complementary Metal-Oxide-Semiconductor}
\acrodef{CMPDP}[C-MPDP]{Calcium-based MPDP}
\acrodef{DPI}[DPI]{Differential Pair Integrator}
\acrodef{DPSS}[DPSS]{Dendritic Prediction of Somatic Spiking}
\acrodef{LTD}[LTD]{Long-Term Depression}
\acrodef{LTP}[LTP]{Long-Term Potentiation}
\acrodef{MPDP}[MPDP]{Membrane Potential Dependent Plasticity}
\acrodef{HMPDP}[H-MPDP]{Homeostatic MPDP}
\acrodef{NMDA}[NMDA]{N-Methyl-D-Aspartate}
\acrodef{OPAMP}[OpAmp]{Operational Amplifier}
\acrodef{OTA}[OTA]{Operational Transconductance Amplifier}
\acrodef{TA}[TA]{Transconductance Amplifier}
\acrodef{RDSP}[RDSP]{Rate Dependent Synaptic Plasticity}
\acrodef{RNN}[RNN]{Recurrent Neural Network}
\acrodef{RTRL}[RTRL]{Real-Time Recurrent Learning}
\acrodef{SDSP}[SDSP]{Spike-Driven Synaptic Plasticity}
\acrodef{SNN}[SNN]{Spiking Neural Network}
\acrodef{SOA}[SOA]{State-of-the-art}
\acrodef{STDP}[STDP]{Spike-Timing Dependent Plasticity}
\acrodef{TSTDP}[T-STDP]{Triplet-based STDP}
\acrodef{VSTDP}[V-STDP]{Voltage-based STDP}
\acrodef{CSTDP}[C-STDP]{Calcium-based STDP}
\acrodef{BDSP}[BDSP]{Burst-Dependent Synaptic Plasticity}
\acrodef{VDCC}[VDCC]{voltage-dependent $Ca^{2+}$ channel}
\acrodef{WTA}[WTA]{Winner-Take-All}
\acrodef{SBCM}[SBCM]{Spiking BCM}

\begin{document}

\topical[Spike-based local synaptic plasticity: A survey of models and circuits]{Spike-based local synaptic plasticity: A survey of computational models and neuromorphic circuits}


\author{Lyes Khacef$^{1, 2}$, Philipp Klein$^{1, 2}$, Matteo Cartiglia$^3$, \\ Arianna Rubino$^3$, Giacomo Indiveri$^3$, Elisabetta Chicca$^{1, 2}$}

\address{$^1$ Bio-Inspired Circuits and Systems (BICS) Lab. Zernike Institute for Advanced Materials, University of Groningen, the Netherlands.}
\address{$^2$ Groningen Cognitive Systems and Materials Center (CogniGron), University of Groningen, the Netherlands. }
\address{$^3$ Institute of Neuroinformatics, University of Zurich and ETH Zurich, Switzerland.}

\eads{\mailto{l.khacef@rug.nl}}
\vspace{10pt}

September 2022


\begin{abstract}
Understanding how biological neural networks carry out learning using spike-based local plasticity mechanisms can lead to the development of powerful, energy-efficient, and adaptive neuromorphic processing systems. A large number of spike-based learning models have recently been proposed following different approaches. However, it is difficult to assess if and how they could be mapped onto neuromorphic hardware, and to compare their features and ease of implementation. To this end, in this survey, we provide a comprehensive overview of representative brain-inspired synaptic plasticity models and mixed-signal \acs{CMOS} neuromorphic circuits within a unified framework. We review historical, bottom-up, and top-down approaches to modeling synaptic plasticity, and we identify computational primitives that can support low-latency and low-power hardware implementations of spike-based learning rules. We provide a common definition of a locality principle based on pre- and post-synaptic neuron information, which we propose as a fundamental requirement for physical implementations of synaptic plasticity. Based on this principle, we compare the properties of these models within the same framework, and describe the mixed-signal electronic circuits that implement their computing primitives, pointing out how these building blocks enable efficient on-chip and online learning in neuromorphic processing systems.

\end{abstract}



\section{Introduction}
\label{sec:intro}
The ability of biological systems to learn and adapt to their environment is key for survival. 
This learning ability is expressed mainly as the change in strength of the synapses that connect neurons, to adapt the structure and function of the underlying network.
The neural substrate of this ability has been studied and modeled intensively, and many brain-inspired learning rules have been proposed~\cite{McNaughton_etal78,Gerstner_etal93,Stuart_Sakmann94,Markram_etal95}.  
The vast majority, if not all, of these biologically plausible learning models rely on local plasticity mechanisms, where locality is a fundamental computational principle, naturally emerging from the physical constraints of the system.
The principle of locality in synaptic plasticity presupposes that all the information a synapse needs to update its state (e.g., its synaptic weight) is directly accessible in space and immediately accessible in time. This information is based on the activity of the pre- and post-synaptic neurons to which the synapse is connected, but not on the activity of other neurons to which the synapse is not physically connected~\cite{Zenke_Neftci21}.

From a biological perspective, locality is a key paradigm of cortical plasticity that supports self-organization, which in turn enables the emergence of consistent representations of the world~\cite{Varela_etal91}. 
From the hardware development perspective, the principle of locality is a key paradigm for the design of spike-based plasticity circuits integrated in embedded systems, in order to enable them to learn online, efficiently and without supervision. 
This is particularly important in recent times, as the rapid growth of wearable and specialized autonomous sensory-processing devices brings new challenges in analysis and classification of sensory signals and streamed data at the edge.
Consequently, there is an increasing need for online learning circuits that have low latency, are low power, and do not need to be trained in a supervised way with large labeled data-sets. 
As standard von Neumann computing architectures have separated processing and memory elements, they are not well suited for simulating parallel neural networks, they are incompatible with the locality principle, and they require a large amount of power compared to in-memory computing architectures. In contrast, neuromorphic architectures typically comprise parallel and distributed arrays of synapses and neurons that can perform computation using only local variables, and can achieve extremely low-energy consumption figures. 
In particular, analog neuromorphic circuits operate the transistors in the weak inversion regime using extremely low currents (ranging from pico-Amperes to micro-Amperes), small voltages (in the range of a few hundreds of milli-Volts), and use the physics of their devices to directly emulate neural dynamics~\cite{Mead90}.
The spike-based learning circuits implemented in these architectures can exploit the precise timing of spikes and consequently take advantage of the high temporal resolutions of event-based sensors. Furthermore, the sparse nature of the spike patterns produced by neuromorphic sensors and processors can give these devices even higher gains in terms of energy efficiency. 

Given the requirements to implement learning mechanisms using limited resources and local signals, animal brains still remain one of our best sources of inspiration, as they have evolved to solve similar problems under similar constraints, adapting to changes in the environment and improving their survival chances~\cite{Hofman15}. 
Bottom-up, brain-inspired approaches to implement learning with local plasticity can be very challenging for solving real-world problems, because of the lack of a clear methodology for choosing specific plasticity rules, and the inability to perform global function optimization (as in gradient back-propagation)~\cite{Eshraghian_etal21}. 
However, these approaches have the potential to support massively parallel and distributed computations and can be used for adaptive online systems at a minimum energy cost~\cite{Neftci_etal19}.
Recent work has explored the potential of brain-inspired self-organizing neural networks with local plasticity mechanisms for spatio-temporal feature extraction~\cite{Bichler_etal12}, unsupervised learning~\cite{Diehl_Cook15,Iyer_Basu17,Hazan_etal18,Kheradpisheh_etal18,Khacef_etal20b}, multi-modal association~\cite{Khacef_etal20,Rathi_Roy21}, adaptive control~\cite{DeWolf_etal20}, and sensory-motor interaction~\cite{Lallee_Dominey13,Zahra_Navarro-Alarcon19}. 

Some of the recently proposed models of plasticity have introduced the notion of a ``third factor'', in addition to the two factors used in learning rules, derived from local information present at the pre- and post-synaptic site.
In these three-factor learning rules, the local variables are used to determine the potential change in the weight (e.g., by using a local eligibility trace), but the change in the weight is applied only when  the additional third factor is presented. This third factor represents a feedback signal (e.g., reward, punishment, or novelty) which could be implemented in the brain for example by diffusion of neuromodulators, such as dopamine~\cite{Kusmierz_etal17,Gerstner_etal18}.
While this feedback signal is locally accessible to the synapse, it is not produced directly at the pre- or post-synaptic site. Therefore, these three-factor learning rules violate the principle of locality that we consider in this review.

In the next section, we provide an overview of synaptic plasticity from a historical, experimental, and theoretical perspective, with a focus on compatibility with physical emulation on \ac{CMOS} systems. 
We then present a selection of representative spike-based synaptic plasticity models that adhere to the principle of locality and that can therefore be implemented in neuromorphic hardware.
We then present analog \ac{CMOS} circuits that implement the basic mechanisms present in the rules discussed. As different implementations have different characteristics that impact the type and number of elements that use local signals, for each target implementation, we assess the principle of locality taking into account the circuits' physical constraints. We conclude proposing steps to reach a unified plasticity framework and presenting the challenges that still remain open in the field.


\section{Synaptic plasticity overview}

\subsection{A brief history of plasticity}
\label{sec:history}
The quest for understanding learning in human beings is a very old one, as the process of acquiring new skills and knowledge was already a subject of debate among philosophers back in Ancient Greece where Aristotle introduced the notion of the brain as a blank state (or \emph{tabula rasa}) at birth that was then developed through education~\cite{Markram_etal11}. It was in contrast to the idea of Plato, his teacher, who believed the brain was pre-formed in the ``heavens'' then sent to earth to join the body. In modern times, the question of nature versus nurture is still being debated, with the view that we are born without preconceptions and our brain is molded by experience proposed by modern philosophers such as~\citeasnoun{Locke89}, and the studies that emphasize the importance of pre-defined structure in the nervous system and in neural networks, to guide and  facilitate the learning process~\cite{Binas_etal15,Hawkins_etal17,Suarez_etal21}.

In the later half of the nineteenth century, learning and memory were linked for the first time to ``junctions between cells'' by~\citeasnoun{Bain73}, even before the discovery of the synapse. In 1890, the psychologist William James postulated a mechanism for associative learning in the brain: ``When two elementary brain-processes have been active together or in immediate succession, one of them, on reoccurring, tends to propagate its excitement into the other''~\cite{James90}. In the same period, neuroanatomists discovered the two main components of the brain: neurons and synapses. They postulated that the brain is composed of separate neurons~\cite{Waldeyer91}, and that long-term memory requires the growth of new connections between existing neurons~\cite{Ramon-y-Cajal94}. These connections became known then as ``synapses''~\cite{Sherrington97}. At the end of the nineteenth century, synapses were already thought to control and change the flow of information in the brain, thus being the substrate of learning and memory~\cite{Markram_etal11}.

The first half of the twentieth century confirmed this hypothesis by various studies on the chemical synapses and the direction of information flow among neurons, going from the pre-synaptic axons to the post-synaptic dendrites. Neural processing was associated to the integration of synaptic inputs in the soma, and the emission of an output spike once a certain threshold was reached, propagating along the axon. Donald Hebb combined earlier ideas and recent discoveries on learning and memory in his book ``The Organization of Behavior''. Similarly to the ideas of James 60 years earlier, Hebb published, in 1949, his formal postulates for the neural mechanisms of learning and memory: ``When an axon of cell A is near enough to excite a cell B and repeatedly or persistently takes part in firing it, some growth process or metabolic change takes place in one or both cells such that A’s efficiency, as one of the cells firing B, is increased''~\cite{Hebb49}. Although Hebb stated that this idea is old, strengthening synapses (that is, increasing synaptic efficacy or weight) connecting co-active neurons has since been called ``Hebbian plasticity''. It is also called \ac{LTP}.

Even though Hebb wrote that ``less strongly established memories would gradually disappear unless reinforced through a slow ``synaptic decay''~\cite{Hebb49}, he did not provide an active mechanism for weakening synapses. Hence, the synaptic strengths or ``weights'' are unbounded and it is not possible to forget previously learned patterns to learn new ones. The first solution proposed a few years later was to maintain the sum of synaptic weights in a neuron constant~\cite{Rochester_etal56}. 
In 1982, Oja proposed a Hebbian-like rule~\cite{Oja82} that adds a ``forgetting'' parameter and solves the stability problem with a form of local multiplicative normalization for synaptic weights. In the same year,~\citeasnoun{Bienenstock_etal82} proposed the \acf{BCM} learning rule where during pre-synaptic stimulation, low-frequency activity of the post-synaptic neuron leads to \ac{LTD} while high-frequency activity would lead to \ac{LTP}. This model was an important shift as it introduced the so-called homo-synaptic \ac{LTD}, where the plasticity was determined by the post-synaptic spike rate with no requirement on the temporal order of spikes. The importance of the post-synaptic neuron in synaptic plasticity was further demonstrated by showing how post-synaptic sub-threshold depolarization can determine whether \ac{LTP} or \ac{LTD} is applied~\cite{Artola_etal90,Sjostrom_etal01}.

Time is inherently present in any associative learning since it only relies on co-occurring events. \citeasnoun{McNaughton_etal78} were the first to experimentally explore the importance of the pre- and post-synaptic spike timing in plasticity. Fifteen years later,~\citeasnoun{Gerstner_etal93} hypothesized that these pre/post spike times contain more information for plasticity compared to spike rates. Their hypothesis would be confirmed by experiments conducted by~\citeasnoun{Stuart_Sakmann94} who discovered that the post-synaptic spike is back-propagating into the dendrites, as well as by~\citeasnoun{Markram_etal95} who showed that a single spike leaves behind a Calcium trace of about \SI{100}{\ms} which is propagated back into the dendrites. These findings were highly influential in the field because they provided evidence that synapses have local access to the timings of pre-synaptic and postsynaptic neurons spikes. In their subsequent experiments,~\citeasnoun{Markram_etal95} provided additional evidence that precise timing is important in neocortical neurons: They showed that using a pre/post pairing with a time difference of \SI{10}{\ms} led to \ac{LTP}, while using the same time difference of \SI{10}{\ms} in an inverted post/pre pairing led to \ac{LTD}~\cite{Markram_etal97}. Larger time differences of \SI{100}{\ms} did not lead to any change in the synaptic weights. Almost concurrently,~\citeasnoun{Bi_Poo98} performed similar experiments and found a \SI{40}{\ms} coincidence time window using paired recordings. These experiments proved that in addition to mean rates, also spike-timing matters. 
This phenomenon was later formulated in a learning rule named \ac{STDP}~\cite{Song_etal00}.

In this respect, the Hebbian learning formula proposed by~\citeasnoun{Shatz92} that ``cells that fire together wire together'' could be misleading, as \possessivecite{Hebb49} postulate is directional: ``axon of cell A is near enough to excite a cell B'', which may be interpreted as implicitly time-dependent since cell A has to fire before cell B. On the other hand, \ac{STDP} had been later found to only partially explain more elaborate learning protocols, which showed that while both \ac{LTP} and \ac{LTD} are compatible \ac{STDP} at low frequencies, only \ac{LTP} occurs at high frequencies regardless of the temporal order of spikes~\cite{Sjostrom_etal01}. 
As pair-based \ac{STDP} models do not reproduce the frequency dependence of synaptic plasticity, \citeasnoun{Pfister_Gerstner06} proposed \acf{TSTDP} rule where \ac{LTP} and \ac{LTD} depend on a combination of three pre- and post-synaptic spikes (either two pre- and one post or one pre- and two post). Both pair-based and triplet-based \ac{STDP} were then shown to be able to reproduce \ac{BCM} like behavior~\cite{Gjorgjieva_etal11}. 
Furthermore, the same frequency dependent experiments~\cite{Sjostrom_etal01} showed that the state of the post-synaptic membrane voltage is important for driving \ac{LTP} or \ac{LTD} under the same pre/post timing conditions, confirming previous studies on the role of the neuron membrane voltage in plasticity~\cite{Artola_etal90}. 
Therefore, these recent findings supported the computational plasticity models that depend on the arrival of the pre-synaptic spike and the voltage of the postsynaptic membrane~\cite{Fusi_etal00,Brader_etal07,Clopath_etal10}, and which were also compatible with the \ac{STDP} model.
The more recent three-factor learning rules aim at bridging the gap between the different time scales of learning, specifically from pre-post spike timings (milliseconds) to behavioral time scales (seconds)~\cite{Gerstner_etal18}.

Today, after more than two millennia of questioning, experimenting and more recently modeling, synaptic plasticity is still not fully understood and many questions remain unanswered. 
Nevertheless, it is clear that multiple forms of plasticity and time-scales co-exist in the synapse and in the whole brain~\cite{Nelson_etal02}. 
They link to each other by sharing locality as a fundamental computational principle. 


\subsection{Experimental perspective}
Synaptic weights are correlated with various elements in biological synapses~\cite{Bartol_etal15b} such as the number of docked vesicles in the pre-synaptic terminal~\cite{Harris_Sultan95}, the area of the pre-synaptic active zone~\cite{Schikorski_etal97}, the dendritic spine head size~\cite{Harris_Stevens89,Hering_Sheng01}, the amount of released transmitters~\cite{Murthy_etal01,Branco_etal08,Ho_etal11}, the area of the post-synaptic density~\cite{Lisman_Harris94}, and the number of AMPA receptors~\cite{Bourne_etal13,Biology20}.
Synaptic plasticity is known to be heterogeneous across different types of synapses~\cite{Abbott_Nelson00,Bi_Poo01}, and there is no unified experimental protocol to confront the different observations.
Here we present the experimental results that led to the bottom-up definition of multiple plasticity rules.

\paragraph{Spike-timing dependence.}
Multiple experiments have been performed to demonstrate the dependence of plasticity on the exact pre- and post-synaptic neurons spike times~\cite{Markram_etal97,Bi_Poo98,Sjostrom_etal01}. From a computational point of view, these experiments led to the proposal of the \ac{STDP} learning rule~\cite{Abbott_Nelson00,Markram_etal11}, and its variants, such as \ac{TSTDP}~\cite{Pfister_Gerstner06}. Typically in these experiments, a pre-synaptic neuron is driven to fire shortly before or shortly after a postsynaptic one, by injecting a current pulse to the specific soma at the desired time. Specifically, these pre-post and post-pre pairings are repeated for \numrange{50}{100} times at a relatively low frequency of about \SIrange{1}{10}{\hertz}~\cite{Sjostrom_Gerstner10}. Experimental results reveal synaptic plasticity mechanisms that are sensitive to the difference in spike times at the time scale of milliseconds~\cite{Gerstner_etal93}. \ac{LTP} is observed when the pre-synaptic spike occurs within \SI{10}{\ms} before the post-synaptic spike is produced, while \ac{LTD} is observed when the order is reversed~\cite{Markram_etal97,Bi_Poo98}. In biology, this precise spike timing dependence could be supported by local processes in the synapses that have access to both the timing information of pre-synaptic spikes and to the postsynaptic spike times, either by sensing their local membrane voltage changes or by receiving large depolarizations caused by output spikes that are back-propagated into the dendrite~\cite{Stuart_Sakmann94}.

\paragraph{Post-synaptic membrane voltage dependence.}
Another feature of synaptic plasticity is its dependence on the post-synaptic neuron membrane voltage~\cite{Artola_etal90}. To study this dependence, the pre-synaptic neuron is driven to fire while the post-synaptic neuron is clamped to a fixed voltage. The clamped voltage level will determine the outcome of the synaptic changes: If the voltage is only slightly above the resting potential of the neuron, then \ac{LTD} is observed while if it is higher, then \ac{LTP} is observed~\cite{Artola_etal90,Ngezahayo_etal00}. These experiments show that post-synaptic spikes are not strictly necessary to induce long-term plasticity~\cite{Lisman_Spruston05,Lisman_Spruston10}. Moreover, even in the presence of a constant pre/post timing (\SI{10}{\ms}) at low frequencies (\SI{0.1}{\hertz}), the post-synaptic membrane voltage determines whether \ac{LTP} or \ac{LTD} can be induced~\cite{Sjostrom_etal01,Sjostrom_Gerstner10}. 
These findings suggest that the post-synaptic membrane voltage might be more important than the pre/post spike timing for synaptic plasticity. 

\paragraph{Frequency dependence.}
While both spike-timing and post-synaptic membrane voltage dependence are observed in experimental protocols when relatively low spike frequencies are used, at high frequencies \ac{LTP} tends to dominate over \ac{LTD} regardless of precise spike timing~\cite{Sjostrom_etal01}. This spike-rate dependence, which is correlated with the Calcium concentration of the postsynaptic neuron~\cite{Sjostrom_etal01}, is captured by multiple learning rules such as \ac{BCM}~\cite{Bienenstock_etal82} or the \ac{TSTDP}~\cite{Pfister_Gerstner06} rule. In these rules, high spike rates produce a strong / rapid increase in Calcium concentration that leads to \ac{LTP}, while low spike rates produce a modest / slow increase in Calcium concentration that decays over time and leads to \ac{LTD}~\cite{Bliss_Collingridge93}. 


\subsection{Theoretical perspective}
Theoretical investigations of plasticity have yielded crucial insights in computational neuroscience. Here, we summarize the fundamental theoretical and practical requirements for long-term synaptic plasticity.


\paragraph{Sensitivity to pre-post spikes correlations.}
Synaptic plasticity has to adjust the synaptic weights depending on the correlation between the pre- and post-synaptic neurons~\cite{Hebb49}. 
Depending on how information is encoded, this can be achieved using spike times, spike rates or both~\cite{Brette15}.
It is important to note that the objective behind the detection of correlation is to detect causality which would ensure a better prediction~\cite{Vigneron_Martinet20}. Even if correlation does not imply causality~\cite{Brette15}, correlation can be considered as a tangible trace for causality in learning.

\paragraph{Selectivity to different patterns.}

In supervised, semi-supervised and reinforcement learning, post-synaptic neurons are driven by a specific teacher signal that forces target neurons to spike and other neurons to remain silent, allowing them to become selective to the pattern applied in input~\cite{Brader_etal07}. In unsupervised learning, the selectivity emerges from competition among neurons~\cite{Kohonen90,Olshausen_Field96} like in \ac{WTA} networks~\cite{Chen17}. 
By associating local plasticity with a \ac{WTA} network, it is possible to create internal models of the probability distributions of the input patterns. This can be interpreted as an approximate Expectation-Maximization algorithm for modeling the input data~\cite{Nessler_etal09}. Recently, the combination of \ac{STDP} with \ac{WTA} networks has been successfully used for solving a variety of pattern recognition problems in both supervised \cite{Chang_etal18} and unsupervised scenarios~\cite{Bichler_etal12,Diehl_Cook15,Iyer_Basu17,Rathi_Roy21}.

\paragraph{Stability of synaptic memory.}
\label{sec:stability}
Long-term plasticity requires continuous adaptation to new patterns but it also requires the retention of previously learned patterns. 
As any physical system has a limited storage capacity, the presentation of new experiences will continuously generate new memories that would eventually lead to saturation of the capacity. When presenting new experiences, the stability (and retrieval) of old memories is a major problem in \acp{ANN}. 
When learning of new patterns leads to the complete corruption or destruction of previously learned ones, then the network undergoes \textit{catastrophic forgetting}~\cite{Nadal_etal86,French99}. 
Both catastrophic forgetting and continual learning are critical problems that need to be addresses for always-on neural processing systems, including artificial embedded processors applied to solving edge-computing tasks. 
The main challenge in always-on learning is not its resilience against time, but its resilience against ongoing activity~\cite{Fusi_etal05}.

Different strategies can be used to find a good balance between plasticity and stability. A first solution is to introduce stochasticity in the learning process, for example by using Poisson distributed spike trains to represent input signals to promote plasticity, while promoting stability using a bi-stable internal variable that slowly drives the weight between one of two possible stable states~\cite{Brader_etal07}. As a result, only a few synapses will undergo a \ac{LTP} or \ac{LTD} transition for a given input, to progressively learn new patterns without forgetting previously learned patterns. 
A second solution is to have an intrinsic stop-learning mechanism to modulate learning and not change synaptic weights if there is enough evidence that the current input pattern has already been learned.


Depending on the particular pattern recognition problem to be solved and the learning paradigm (offline/online), specific properties can be more or less important.


\section{Computational primitives of synaptic plasticity}
In this work, we refer to ``computational primitives of synaptic plasticity'' as those basic plasticity mechanisms that make use of local variables.

\subsection{Local variables}
\label{sec:local-var}

\begin{center}
    \begin{figure}[H]
        \includegraphics[width=\textwidth, angle = 0 ]{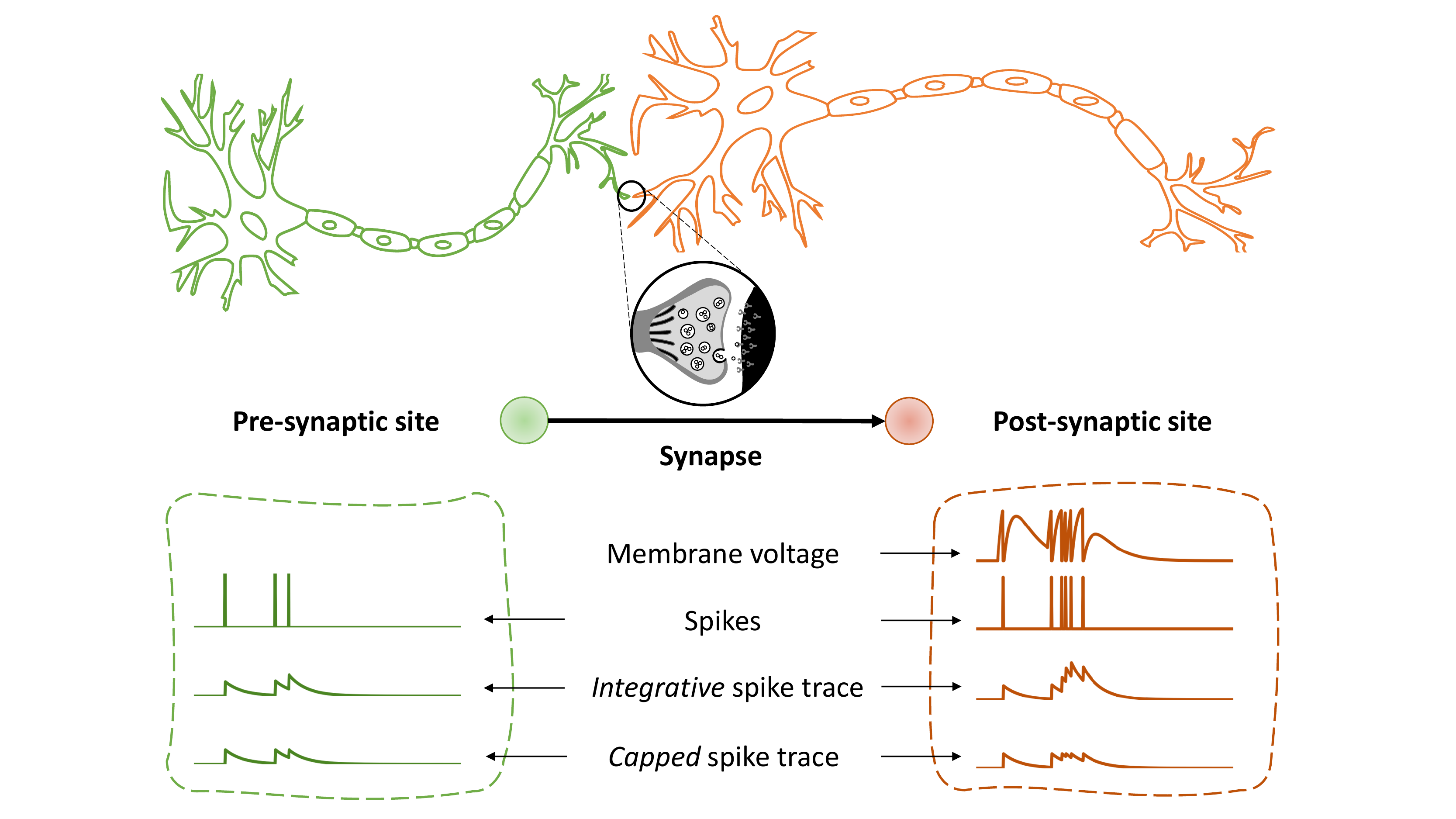}
        \centering
        \caption{The local variables involved in the local synaptic plasticity models we review in this survey: Pre- and/or post-synaptic spike traces (capped or integrative) and post-synaptic membrane (dendritic or somatic) voltage.}
        \label{fig:local_var}
    \end{figure}
\end{center}

The following are the local variables that we consider:

\begin{description}
\item[Pre- and post-synaptic spike traces:]
  These are the traces generated at the pre- and post-synaptic site triggered by the spikes of the corresponding pre- or post-synaptic neurons.
  They can be computed by either integrating the spikes using a linear operator in models and a low-pass filter in circuits, or by using non-linear operators/circuits. Figure~\ref{fig:local_var} shows examples both linear (denoted as ``integrative'') and non-linear (denoted as ``capped'') spike traces.
  In general, these traces represent the recent level of activation of the pre- and post-synaptic neurons. 
  Depending on the learning rule, there might be one or more spike traces per neuron with different decay rates.
  The biophysical substrates of these traces can be diverse~\cite{Pfister_Gerstner06,Graupner_Brunel10}, for example reflecting the amount of bound glutamate~\cite{Karmarkar_Buonomano02} or the number of \ac{NMDA} receptors in an activated state~\cite{Senn_etal01}. The post-synaptic spike traces could reflect the Calcium concentration mediated through voltage-gated Calcium channels and \ac{NMDA} channels~\cite{Karmarkar_Buonomano02}, the number of secondary messengers in a deactivated state of the \ac{NMDA} receptor~\cite{Senn_etal01} or the voltage trace of a back-propagating action potential~\cite{Shouval_etal02}.
  
\item[Post-synaptic membrane voltage:]  
  The post-synaptic neuron's membrane potential is also a local variable, as it is accessible to all of the neuron's synapses. 
\end{description}

These local variables are the basic elements that can be used to induce a change in the synaptic weight, which is reflected in the change of the post-synaptic membrane voltage that a pre-synaptic spike induces.


\subsection{Spikes interaction}
\label{sec:spike-interaction}
We refer to spike interactions as the number of spikes from the past activity of the neurons that are taken into account for the weight update. In particular, we distinguish two spikes interaction schemes:

\begin{description}
    \item[All-to-all:] In this scheme, the spike trace is "integrative" and influenced, asymptotically, by the whole previous spiking history of the pre-synaptic neuron. The contribution of each spike is expressed in the form of a Dirac delta which should be integrated. Nevertheless, if the spikes are considered to be point processes for which their spike width is zero in the limit, the contribution of all spikes in  Eq.~\eqref{eq:trace-rate} can be approximated as follows:
    
    \begin{equation}
    \centering
    \label{eq:trace-rate}
         \frac{dX(t)}{dt} = - \frac{X(t)}{\tau} + \sum _{i} A \: \delta \left ( t - t_i \right )
    \end{equation}
    
    where $\delta \left ( t - t_i \right )$ is a spike occurring at time $t_i$, $\tau$ is the exponential decay time constant and $A$ is the jump value such that at the moment of a spike event, \textit{the spike trace jumps by $A$}. In addition to being a good first-order model of synaptic transmission, this transfer function can be easily implemented in electronic hardware using low-pass filters. Indeed, the trace $X(t)$ represents the online estimate of the neuron's mean firing rate~\cite{Dayan_Abbott01}.
    
    \item[Nearest spike:] This is a non-linear mode in which the spike trace is only influenced by the most recent pre-synaptic spike. It is implemented by means of a hard bound that is limiting the maximum value of the trace, such that if the jumps reach it, the trace is "capped" at that bound value. It is expressed in Eq.~\eqref{eq:trace-time}:
    
    \begin{equation}
    \centering
    \label{eq:trace-time}
        \frac{dX(t)}{dt} = - \frac{X(t)}{\tau} + \sum _{i} (A - X(t)) \: \delta \left ( t - t_i \right )
    \end{equation}
    
    where $A$ is both the jump value and the hard bound, such that at the moment of a spike event, \textit{the spike trace jumps to $A$}. It means that the spike trace gives an online estimate of the time since the last spike.
\end{description}

Therefore, the jump and bound parameters control the sensitivity of the learning rule to the spike timing and rate combined (all-to-all) or to the spike timing alone (nearest spike), while the decay time constant controls how fast the synapse forgets about these activities. 
Further spike interaction schemes are possible, for example by adapting the nearest spike interaction so that spike interactions producing \ac{LTP} would dominate over those producing \ac{LTD}.

\subsection{Update trigger}
In most synaptic plasticity rules, the weights update is event-based and happens at the moment of a pre-synaptic spike~\citeaffixed{Brader_etal07}{e.g.}, post-synaptic spike~\citeaffixed{Diehl_Cook15}{e.g.} or both pre- and post-synaptic spikes~\citeaffixed{Song_etal00}{e.g.}. This event-based paradigm is particularly interesting for hardware implementations, as it exploits the spatio-temporal sparsity of the spiking activity to reduce the energy consumption with less updates. On the other hand, some rules use a continuous update~\citeaffixed{Graupner_Brunel12}{e.g.} arguing for more biological plausibility, or a mixture of both with e.g.\ depression at the moment of a pre-synaptic spike and continuous potentiation~\citeaffixed{Clopath_etal10}{e.g.}.


\subsection{Synaptic weights}
The synaptic weight represents the strength of a connection between two neurons. 
Synaptic weights have three main characteristics:

\begin{enumerate}
    \item Type: Synaptic weights can be continuous, with full floating-point resolution in software, or with fixed/limited resolution (binary in the extreme case). Both cases can be combined by using fixed resolution synapses (e.g., binary synapses), which however have a continuous internal variable that determines if and when the synapse undergoes a low-to-high (\ac{LTP}) or high-to-low (\ac{LTD}) transition, depending on the learning rule.
    
    \item Bistability: In parallel to the plastic changes that update the weights, on their weight update trigger conditions, synaptic weights can be continuously driven to one of two stable states, depending on additional conditions on the weight itself and on its recent history. These bistability mechanisms have been shown to protect memories against unwanted modifications induced by ongoing spontaneous activity~\cite{Brader_etal07} and provide a way to implement stochastic selection mechanisms.
    
    \item Bounds: In any physical neural processing system, whether biological or artificial, synaptic weights have bounds: they cannot grow to infinity. Two types of bounds can be imposed on the weights: (1) hard bounds, in rules with additive updates independent of weight, or (2) soft bounds, in weight-dependent updates (for example, multiplicative) rules that drive the weights toward the bounds asymptotically~\cite{Morrison_etal08}.
\end{enumerate}


\subsection{Stop-learning}
\label{sec:stop-learning}
An intrinsic mechanism to modulate learning and automatically switch from the training mode to the inference mode is important, especially in an online learning context. 
This ``stop-learning'' mechanism can be either implemented with a global signal related to the performance of the system, as in reinforcement learning, or with a local signal produced in the synapses or in the soma. 
For example, a local variable that can be used to implement stop-learning could be derived from the post-synaptic neuron's membrane voltage~\cite{Clopath_etal10,Albers_etal16} or spiking activity~\cite{Brader_etal07,Graupner_Brunel12}. 


\section{Models of synaptic plasticity}
\label{sec:models}
We present a representative set of spike-based synaptic plasticity models,  summarize their main features, and explain their working principles. Table~\ref{tab:models} shows a direct comparison of the computational principles used by the relevant models, and Tables~\ref{tab:models-variablesI} and~\ref{tab:models-variablesII} show the main variables common to the different models. 

\begin{center}
    \begin{table}[H]
        \caption{Spike-based local synaptic plasticity rules: comparative table}
        \label{tab:models}
        \resizebox{\textwidth}{!}{%
        \begin{tabular}{>{\hspace{0pt}}m{0.1\linewidth}>{\centering\hspace{0pt}}m{0.27\linewidth}>{\centering\hspace{0pt}}m{0.09\linewidth}>{\centering\hspace{0pt}}m{0.08\linewidth}>{\centering\hspace{0pt}}m{0.09\linewidth}>{\centering\hspace{0pt}}m{0.1\linewidth}>{\centering\hspace{0pt}}m{0.08\linewidth}>{\centering\hspace{0pt}}m{0.09\linewidth}>{\centering\arraybackslash\hspace{0pt}}m{0.1\linewidth}} 
        \hline
        \multirow{2}{\linewidth}{\hspace{0pt}\textbf{Plasticity rule}} & \multirow{2}{\linewidth}{\hspace{0pt}\Centering{}\textbf{Local variables}} & \multirow{2}{\linewidth}{\hspace{0pt}\Centering{}\textbf{Spikes interaction}} & \multicolumn{2}{>{\Centering\hspace{0pt}}m{0.2\linewidth}}{\textbf{Update trigger (spike)}} & \multicolumn{3}{>{\Centering\hspace{0pt}}m{0.27\linewidth}}{\textbf{Synaptic weights}} & \multirow{2}{\linewidth}{\hspace{0pt}\Centering{}\textbf{Stop-learning}} \cr 
        \cline{4-8}
         &  &  & \textbf{\acs{LTD}} & \textbf{\acs{LTP}} & \textbf{Type} & \textbf{Bistability} & \textbf{Bounds} &  \cr 
        \hline
        \textbf{\acs{STDP}} & Pre- and post-synaptic spike traces & Nearest spike & Pre & Post & Analog & No & Hard & No \cr 
        \hline
        \textbf{\acs{TSTDP}} & Pre-synaptic spike trace + 2 post-synaptic spike traces (different time constants) & Nearest spike / all-to-all & Pre & Post & Analog & No & Hard & No \cr 
        \hline
        \textbf{\acs{SDSP}} & Post-synaptic membrane voltage + post-synaptic spike trace & All-to-all & \multicolumn{2}{>{\Centering\hspace{0pt}}m{0.2\linewidth}}{Pre} & Binary$^*$ & Yes & Hard & Yes$^1$ \cr 
        \hline
        \textbf{\acs{VSTDP}} & Pre-synaptic spike trace + post-synaptic membrane voltage + 2 post-synaptic membrane voltage traces & All-to-all & Pre & Continuous & Analog & No & Hard & Yes$^2$ \cr 
        \hline
        \textbf{\acs{CSTDP}} & One synaptic spike trace updated by both pre- and post-synaptic spikes & All-to-all & \multicolumn{2}{>{\Centering\hspace{0pt}}m{0.2\linewidth}}{Continuous} & Analog & Yes & Soft & Yes$^3$ \cr 
        \hline
        \textbf{\acs{SBCM}} & Pre- and post-synaptic spike traces & All-to-all & \multicolumn{2}{>{\Centering\hspace{0pt}}m{0.2\linewidth}}{Continuous} & Analog & No & Hard & No \cr 
        \hline
        \textbf{\acs{MPDP}} & Pre-synaptic spike trace + post-synaptic membrane voltage & All-to-all & \multicolumn{2}{>{\Centering\hspace{0pt}}m{0.2\linewidth}}{Continuous} & Analog & No & Hard & Yes$^4$ \cr 
        \hline
        \textbf{\acs{DPSS}} & Pre-synaptic spike trace + post-synaptic dendritic voltage + post-synaptic somatic spike & All-to-all & \multicolumn{2}{>{\Centering\hspace{0pt}}m{0.2\linewidth}}{Continuous} & Analog & No & Hard & No \cr 
        \hline
        \textbf{\acs{RDSP}} & Pre-synaptic spike trace & All-to-all & \multicolumn{2}{>{\Centering\hspace{0pt}}m{0.2\linewidth}}{Post} & Analog & No & Soft & No \cr 
        \hline
        \textbf{\acs{HMPDP}} & Pre-synaptic spike trace + post-synaptic membrane voltage & All-to-all & \multicolumn{2}{>{\Centering\hspace{0pt}}m{0.2\linewidth}}{Continuous} & Analog & No & Hard & Yes$^5$ \cr 
        \hline
        \textbf{\acs{CMPDP}} & Post-synaptic membrane voltage + post-synaptic spike trace & All-to-all & \multicolumn{2}{>{\Centering\hspace{0pt}}m{0.2\linewidth}}{Pre} & Analog & No & Hard & No \cr 
        \hline
        \textbf{\acs{BDSP}} & Pre-synaptic spike trace + post-synaptic event trace + post-synaptic burst trace & All-to-all & Post (event) & Post (burst) & Analog & No & Hard & No \cr
        \hline
        \end{tabular}%
        }
    
        \footnotesize{$^*$ Binary with analog internal variable. \newline
        $^1$ At low and high activities of post-neuron (post-synaptic spike trace). \newline
        $^2$ At low low-pass filtered post-synaptic membrane voltage (post-synaptic membrane voltage trace). \newline
        $^3$ At low activity of pre- and post-neurons merged (synaptic spike trace). \newline
        $^4$ At medium (between two thresholds) internal update trace. \newline
        $^5$ At medium (between two thresholds) post-synaptic membrane voltage.}
    \end{table}
\end{center}


\subsection{Song et al. (2000): \acf{STDP}}
\acf{STDP}~\cite{Song_etal00} was proposed to model how pairs of pre-post spikes interact based solely on their timing. It is one of the most widely used synaptic plasticity algorithms in the literature.

\begin{equation}
    \label{eq:stdp}
    \Delta w = 
    \begin{cases}
        \mathrm{A}_{+} \exp (\frac{\Delta t}{\tau_{+}}) & \text{if $\Delta\mathrm{t}<0$}.\\
        -\mathrm{A}_{-} \exp (\frac{-\Delta t}{\tau_{-}}) & \text{if $\Delta\mathrm{t}\geq 0$}.
    \end{cases}
\end{equation}

\begin{center}
    \begin{figure}[H]
        \includegraphics[width=\textwidth, angle = 0 ]{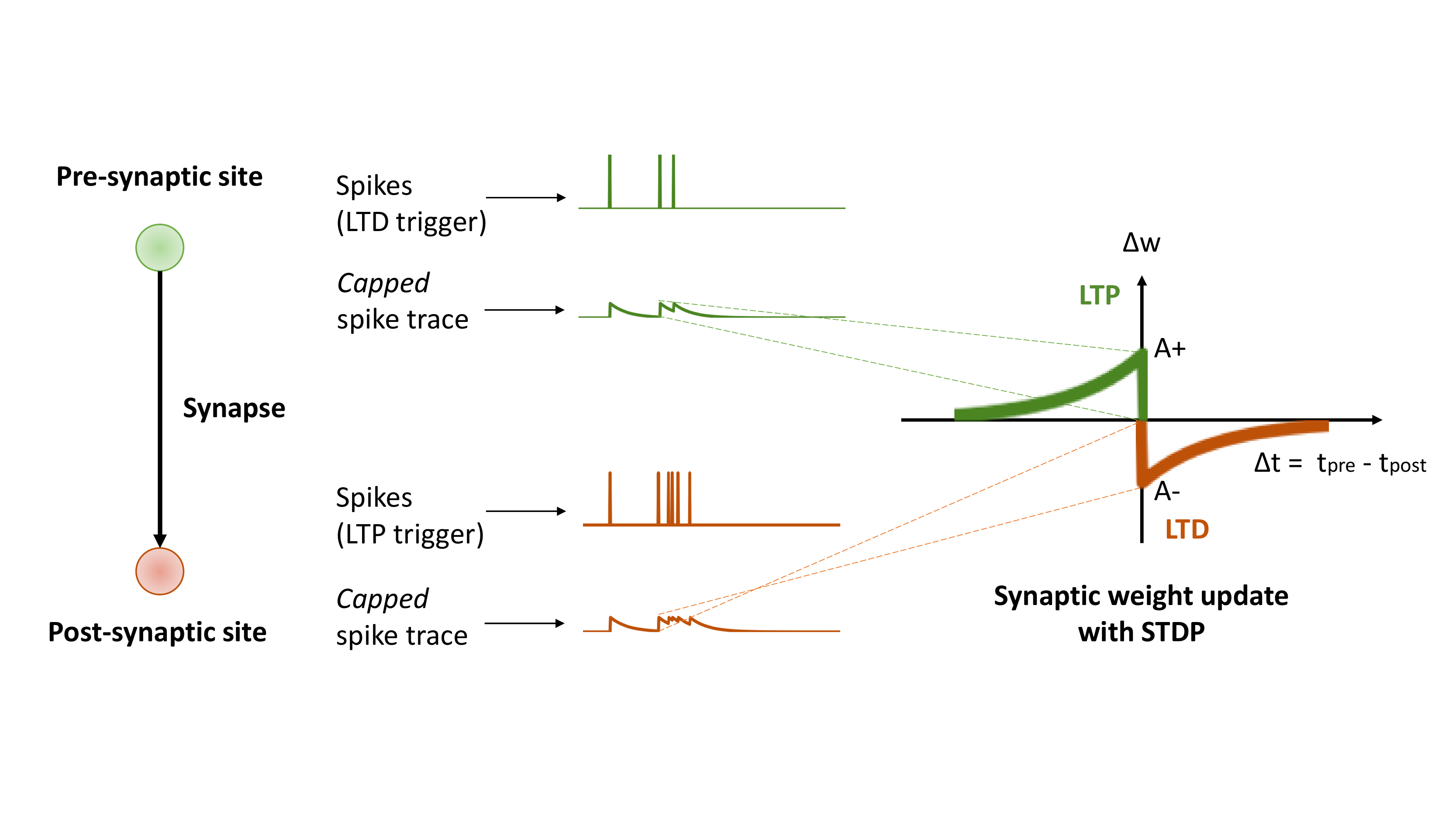}
        \centering
        \caption{Online implementation principle of STDP using local pre- and post-synaptic capped spike traces which provide an online estimate of the time since the last spike. For example, at the moment of post-synaptic spike, potentiation is induced with a weight change that is proportional to the value of the pre-synaptic spike trace, and the post-synaptic spike trace is updated with a jump to $A_{-}$.}
        \label{fig:stdp_traces}
    \end{figure}
\end{center}

The synaptic weight is updated according to Eq.~\eqref{eq:stdp}, whose variables are described in Tab.~\ref{tab:stdp}. If a post-synaptic spike occurs after a pre-synaptic one ($\Delta\mathrm{t}<0$), potentiation is induced (triggered by the post-synaptic spike). In contrast, if a pre-synaptic spike occurs after a post-synaptic spike ($\Delta\mathrm{t}\geq 0$), depression occurs (triggered by the pre-synaptic spike). The time constants $\tau_{+}$ and $\tau_{-}$ determine the time window in which the spike interaction leads to changes in synaptic weight.
As shown in Tab.~\ref{tab:models}, \ac{STDP} is based on local pre- and post-spike traces with nearest spike interaction, meaning that the spike traces are capped. Fig.~\ref{fig:stdp_traces} illustrates how \ac{STDP} is implemented using these spike traces for online learning.

\begin{table}[H]
\centering
    \caption{Variables of the \ac{STDP} rule.}
    \label{tab:stdp}
    \begin{tabular}{@{}cc@{}}
    \toprule
    \textbf{Variable} & \textbf{Description} \cr
    \midrule
    $w$& Synaptic weight\cr
    $\mathrm{A}_{+}$ / $\mathrm{A}_{-} $& Maximum amount of synaptic change\cr
    $\Delta t$& Time difference between pre- and post-synaptic spikes: $t_{pre} - t_{post}$\cr
    $\tau_{+}$ / $\tau_{-}$& Time constants of synaptic traces\cr
    \bottomrule
    \end{tabular}
\end{table}


\subsection{Pfister and Gerstner (2006): \acf{TSTDP}}
The main limitation of the original \ac{STDP} model is that it is only time-based; thus, it cannot reproduce frequency effects as well as triplet and quadruplet experiments. In this work,~\citeasnoun{Pfister_Gerstner06} introduces additional terms in the learning rule to expand the classical pair-based \ac{STDP} to a \acf{TSTDP}.

Specifically, the authors introduce a triplet depression (i.e.\ 2-pre and 1-post) and potentiation term (i.e.\ 1-pre and 2-post). They do this by adding 
four additional variables that they call detectors: $r$ and $o$. $r_{1}$ and $r_{2}$ detectors are pre-synaptic spike traces which increase whenever there is a pre-synaptic spike and decrease back to zero with their individual intrinsic time constants. Similarly, $o_{1}$ and $o_{2}$ detectors increase on post-synaptic spikes and decrease back to zero with their individual intrinsic time constants. The weight changes are defined in Eqs.~\eqref{eq:tstdp}, whose variables are described in Tab.~\ref{tab:tstdp}.

\begin{equation}
\label{eq:tstdp}
    \begin{array}{l}
    w(t)\rightarrow w(t)+r_{1}(t)\left[A_{2}^{+}+A_{3}^{+} o_{2}(t-\epsilon)\right] \text { if } t=t^{\mathrm{{post}}} \\
    w(t)\rightarrow w(t)-o_{1}(t)\left[A_{2}^{-}+A_{3}^{-} r_{2}(t-\epsilon)\right] \text { if } t=t^{\mathrm{pre}}
    \end{array}
\end{equation}

While in classical \ac{STDP}, potentiation takes place shortly after a pre-synaptic spike and upon occurrence of a post-synaptic spike, in the current framework several conditions need to be considered. Potentiation is triggered at every post-synaptic spike where the weight change is gated by the $r_{1}$ detector and modulated by the $o_{2}$ detector. If there are no post-synaptic spikes shortly before the current one ($o_{2}$ is zero) the degree of potentiation is determined by $A_{2}^{+}$ only, just like in the pair-based \ac{STDP}. If however a triplet of spikes occurs (in this case 1-pre and 2-post) $o_{2}$ is non zero and an additional potentiation term $A_{3}^{+} o_{2}(t-\epsilon)$ contributes to the weight change. Analogously, $r_{2}$, $o_{1}$, $A_{2}^{-}$ and $A_{3}^{-}$ operate for the case of synaptic depression which is triggered at every pre-synaptic spike.

\begin{table}[H]
\centering
    \caption{Variables of the \ac{TSTDP} rule.}
    \label{tab:tstdp}
    \begin{tabular}{@{}cc@{}}
    \toprule
    \textbf{Variable} & \textbf{Description} \cr
    \midrule
    $w$& Synaptic weight\cr
    $r_{1}$ / $r_{2}$& Pre-synaptic spike traces - integrative \cr
    $o_{1}$ / $o_{2}$ & Post-synaptic spike traces - integrative\cr
    $\mathrm{A}_{2}^{+}$ / $\mathrm{A}_{2}^{-}$ &Weight change amplitude whenever there is a pair event \cr 
    $\mathrm{A}_{3}^{+}$ / $\mathrm{A}_{3}^{-}$& Weight change amplitude whenever there is triplet event\cr
    $ \epsilon $ & Small positive constant\cr
    $t^{\mathrm{pre}}$ / $t^{\mathrm{post}}$& Time of pre- and post-synaptic spikes\cr
    \bottomrule
    \end{tabular}
\end{table}


\subsection{Brader et al. (2007): \acf{SDSP}}
The \acf{SDSP} learning rule addresses in particular the problem of memory maintenance and catastrophic forgetting: the presentation of new experiences continuously generates new memories that will eventually lead to saturation of the limited storage capacity, hence forgetting.
As discussed in Sec.~\ref{sec:stability}, this problem concerns all learning rules in an online context.
\ac{SDSP} attempts to solve it by slowing the learning process in an unbiased way. The model randomly selects the synaptic changes that will be consolidated among those triggered by the input, therefore learning to represent the statistics of the incoming stimuli.

The \ac{SDSP} model proposed by~\citeasnoun{Brader_etal07} is demonstrated in a feed-forward neural network used for supervised learning in the context of pattern classification. Nevertheless, the model is also well suited for unsupervised learning of patterns of activation in attractor neural networks~\cite{Del-Giudice_etal03,Brader_etal07}.
It does not rely on the precise timing difference between pre- and post-synaptic spikes, instead the weight update is triggered by single pre-synaptic spikes. The sign of the weight update is determined by the post-synaptic neuron's membrane voltage $V(t^{pre})$. The post-synaptic neuron's Calcium variable $C(t^{pre})$ represents a trace of the recent low-pass filtered post-synaptic activity and is used to determine if synaptic updates should occur (stop-learning mechanism). The synaptic dynamics is described in Eq.~\eqref{eq:trace-rate}. 

The internal variable $X$ is updated according to Eq.~\eqref{eq:sdsp} with the variables described in Tab.~\ref{tab:sdsp}.

\begin{equation} 
    \label{eq:sdsp}
    \begin{array}{l}
    X \rightarrow X + a 
    \text{ if } V(t^{\mathrm{pre}}) > \theta_{V} \text { and } \theta_{\mathrm{up}}^{\mathrm{l}} < C(t^{\mathrm{pre}}) <\theta_{\mathrm{up}}^{\mathrm{h}}\\ \\
    X \rightarrow X - b 
    \text{ if } V(t^{\mathrm{pre}}) \leq \theta_{V} \text { and } \theta_{\mathrm{down}}^{\mathrm{l}} < C(t^{\mathrm{pre}}) <\theta_{\mathrm{down}}^{\mathrm{h}}\\
    \end{array}
\end{equation}

The weight update depends on the instantaneous values of $V(t^{\mathrm{pre}})$ and $C(t^{\mathrm{pre}})$ at the arrival of a pre-synaptic spike. A change of the synaptic weight is triggered by the pre-synaptic spike if $V(t^{\mathrm{pre}})$ is above a threshold $\theta_{v}$, provided that the post-synaptic Calcium trace $C(t^{\mathrm{pre}})$ is between the potentiation thresholds $\theta_{up}^{\mathrm{l}}$ and $\theta_{up}^{\mathrm{h}}$. An analogous but flipped mechanism induces a decrease in the weights.

The synaptic weight is restricted to the interval $0 \leq X \leq X_{max}$. The bistability on the synaptic weight implies that the internal variable $X$ drifts (and is bounded) to either a low state or a high state, depending on whether $X$ is below or above a threshold $\theta_{X}$ respectively. This is shown in Eqs~\eqref{eq:sdsp-bistability}. 

\begin{equation} 
    \label{eq:sdsp-bistability}
    \frac{dX}{dt} =
    \begin{cases}
        \alpha & \text {if $X > \theta_{X}$}\\
        -\beta & \text {if $X \leq \theta_{X}$}
    \end{cases}
\end{equation}

\begin{table}[H]
\centering
    \caption{Variables of the \ac{SDSP} rule.}
    \label{tab:sdsp}
    \begin{tabular}{@{}cc@{}}
    \toprule
    \textbf{Variable} & \textbf{Description} \cr
    \midrule
    $X$& Synaptic weight\cr
    $a,b$ & Jump sizes\cr
    $V(t)$& Post synaptic membrane potential\cr
    $\theta_{V}$ & Membrane potential threshold \cr
    $C(t)$& Post-synaptic spike trace (Calcium) - integrative\cr	
    $\theta_{\mathrm{up}}^{\mathrm{l}}$ / $\theta_{\mathrm{up}}^{\mathrm{h}}$ / $\theta_{\mathrm{down}}^{\mathrm{l}}$ / $\theta_{\mathrm{down}}^{\mathrm{h}}$ &Thresholds on the Calcium variable\cr
    $X_{max}$& Maximum synaptic weight\cr
    $\alpha$ / $\beta$ & Bistability rates, $\in\mathbb{R}^+$ \cr
    $\theta_{X}$& Bistability threshold on the synaptic weight\cr
    \bottomrule
    \end{tabular}
\end{table}


\subsection{Clopath et al. (2010): \acf{VSTDP}}
The \acf{VSTDP} rule has been introduced to unify several experimental observations such as post-synaptic membrane voltage dependence, pre-post spike timing dependence and post-synaptic rate dependence~\cite{Clopath_Gerstner10}, but also to explain the emergence of some connectivity patterns in the cerebral cortex~\cite{Clopath_etal10}.
In this model, depression and potentiation are two independent mechanisms whose sum produces the total synaptic change. Variables of the equations are described in Tab.~\ref{tab:vstdp}.

Depression is triggered by the arrival of a pre-synaptic spike ($X(t)=1$) and is induced if the voltage trace $\overline{u}_{-}(t)$ of the post-synaptic membrane voltage $u(t)$ is above the threshold $\theta_{-}$ (see Eq.~\eqref{eq:vstdp_ltd}).

\begin{equation}
\label{eq:vstdp_ltd}
\frac{dw^{-}}{dt} = -A_{\mathrm{LTD}} X(t)[\overline{u}_{-}(t) - \theta_{-}]_{+}
\end{equation}



On the other hand, potentiation is continuous and occurs following Eq.~\eqref{eq:vstdp_ltp} if the following conditions are met at the same time:

\begin{itemize}
    \item The instantaneous post-synaptic membrane voltage $u(t)$ is above the threshold $\theta_{+}$, with $\theta_{+} > \theta_{-}$;
    \item The low-pass filtered post-synaptic membrane voltage $\overline{u}_{+}$ is above $\theta_{-}$;
    \item A pre-synaptic spike occurred a few milliseconds earlier and has left a trace $\overline{x}$.
\end{itemize}

\begin{equation}
\label{eq:vstdp_ltp}
    \frac{dw^+}{dt} = +A_{\mathrm{LTP}}\: \overline{x}(t)\: \left [ u(t) - \theta_{+} \right ]_{+}\: \left [ \overline{u}_{+}(t) - \theta_{-} \right ]_{+}
\end{equation}



The total synaptic change is the sum of depression and potentiation expressed in Eqs.~\eqref{eq:vstdp_ltd} and \eqref{eq:vstdp_ltp} respectively, within the weights' hard bounds $0$ and $w_{\mathrm{max}}$.

\begin{table}[H]
\centering
    \caption{Variables of the \ac{VSTDP} rule.}
    \label{tab:vstdp}
    \begin{tabular}{@{}cc@{}}
    \toprule
    \textbf{Variable} & \textbf{Description} \cr
    \midrule
    $w$ & Synaptic weight\cr
    $X(t)$& Pre-synaptic spike train\cr
     & $X(t) = \sum _{n} \delta \left ( t - t^{n} \right )$\cr
    $\delta(.)$ & Dirac delta function\cr
    $t^{n}$& Time of the n-th pre-synaptic spike\cr
    $u(t)$& Post-synaptic membrane voltage\cr
    $\overline{u}_{-}(t)$ / $\overline{u}_{+}(t)$& Post-synaptic membrane voltage traces \cr
    $A_{\mathrm{LTD}}$ / $A_{\mathrm{LTP}}$ & Amplitudes for depression and potentiation\cr
    $\theta_{-}$ / $\theta_{+}$& Thresholds\cr
    $[.]_{+}$& Rectifying bracket $[x]_+ = x$ if $x>0$, $[x]_+ =0$ otherwise\cr
    $\overline{x}(t)$& Pre-synaptic spike trace - integrative\cr
    $w_{\mathrm{max}}$& Weight max hard bound\cr
    \bottomrule
    \end{tabular}
\end{table}


\subsection{Graupner and Brunel (2012): \acf{CSTDP}}
Founded on molecular studies,~\citeasnoun{Graupner_Brunel12} proposed a plasticity model (\ac{CSTDP}) based on a transient Calcium signal. They model a single Calcium trace variable $c(t)$ which represents the linear sum of individual Calcium transients elicited by pre- and post-synaptic spikes at times $t_i$ and $t_j$, respectively. The amplitudes of the transients elicited by pre- and post-synaptic spikes are given by $C_{\mathrm{pre}}$ and $C_{\mathrm{post}}$, respectively, and $c(t)$ decays constantly towards $0$.


In the proposed model, the synaptic strength is described by the synaptic efficacy $\rho\in[0:1]$, which is constantly updated according to Eq.~\eqref{eq:cstdp}, whose variables are described in Tab.~\ref{tab:cstdp}. 
Changes in synaptic efficacy are continuous and depend on the relative times in which the Calcium trace $c(t)$ is above the potentiation ($\theta_p$) and depression ($\theta_d$) thresholds~\cite{Graupner_Brunel12}.

\begin{equation}
    \begin{split}
        \tau \frac{d\rho}{dt} = -\rho(1 - \rho)(\rho_{\star} - \rho) + \gamma_{p}(1 - \rho)\Theta[c(t) - \theta_p] - \gamma_d \rho \Theta[c(t) - \theta_d] + \mathrm{Noise(t)}
    \end{split}
\label{eq:cstdp}
\end{equation}

If the Calcium variable is above the threshold for potentiation ($\Theta[c(t) - \theta_p] = 1$) the synaptic efficacy is continuously increased by $\frac{\gamma_p(1 - \rho)}{\tau}$ and as long as the Calcium variable is above the threshold for depression ($\Theta[c(t) - \theta_d] = 1$) the synaptic efficacy is continuously decreased by $-\frac{\gamma_d\rho}{\tau}$.
Eventually, the efficacy updates induced by the Calcium concentration are in direct competition with each other as long as $c(t)$ is above both thresholds~\cite{Graupner_Brunel12}. 
In addition to constant potentiation or depression updates, the bistability mechanism $-\rho(1 - \rho)(\rho_{\star} - \rho)$ drives the synaptic strength toward $0$ or $1$, depending on whether the instantaneous value of $\rho$ is below or above the bistability threshold $\rho_{\star}$.

\citeasnoun{Graupner_Brunel12} show that their rule replicates a plethora of dynamics found in numerous experiments, including pair-based behavior \ac{STDP} with different \ac{STDP} curves, synaptic dynamics found in CA3-CA1 slices for postsynaptic neuron spikes and dynamics based on spike triplets or quadruplets. 
However, the rule contains only a single Calcium trace variable $c(t)$ per synapse, which is updated by both pre- and post-synaptic spikes. Since the synaptic efficacy update only depends on this variable and not on the individual or paired spike events of the pre- and post-synaptic neuron, the system can get into a state in which isolated pre-synaptic or isolated post-synaptic activity can lead to synaptic efficacy changes. In extreme cases, isolated pre(post)-synaptic spikes could drive a highly depressed ($\rho = 0$) synapse into the potentiated state ($\rho = 1$), without the occurrence of any post(pre)-synaptic action potential.
In a recent work, \citeasnoun{Chindemi_etal22} uses a modified version of the \ac{CSTDP} rule based on data-constrained post-synaptic Calcium dynamics according to experimental data. They show that the rule is able to replicate the connectivity of pyramidal cells in the neocortex, by adapting the probabilistic and limited release of $Ca^{2+}$ during pre- and post-synaptic activity.

\begin{table}[H]
\centering
    \caption{Variables of the \ac{CSTDP} rule.}
    \label{tab:cstdp}
    \begin{tabular}{@{}cc@{}}
    \toprule
    \textbf{Variable} & \textbf{Description} \cr
    \midrule
    $c(t)$& Pre- and post-synaptic spike trace (Calcium) - integrative\cr
    $C_{\mathrm{pre}}$ / $C_{\mathrm{post}}$& Amplitudes of pre- and post-synaptic Calcium jumps \cr
    $\theta_p$ / $\theta_d$& Thresholds for potentiation and depression\cr
    $\tau$& Time constant of synaptic efficacy changes\cr
    $\rho$& Synaptic efficacy\cr
    $\rho_{\star}$ & Bistability threshold on the synaptic efficacy\cr
    $\gamma_{p}$ / $\gamma_{d}$ & Rates of synaptic potentiation and depression\cr
    $\Theta[.]$ & Heaviside function $\Theta[x] = 1$ if $x>0$, $\Theta[x] =0$ otherwise\cr
    $\mathrm{Noise(t)}$ & Activity-dependent noise\cr
    \bottomrule
    \end{tabular}
\end{table}


\subsection{Bekolay et al. (2013): \acf{SBCM}}
The \acf{SBCM} learning rule~\cite{Bekolay_etal13} has been proposed as another spike-based formulation of the abstract learning rule \ac{BCM}, after the \ac{TSTDP} rule. The weight update of the \ac{SBCM} learning rule is continuous and is expressed in Eq.~\eqref{eq:sbcm}, whose variables are described in Tab.~\ref{tab:sbcm}.

\begin{equation}
\label{eq:sbcm}
    \Delta w_{ij} = \kappa \alpha_j a_i a_j (a_j - \theta(t))
\end{equation}

The mechanistic properties of \ac{SBCM} are closer to the formal \ac{BCM} rule, with the activities of the neurons expressed as spike activity traces and a filtered modification threshold. Nevertheless, the \ac{SBCM} exhibits both the timing dependence of \ac{STDP} and the frequency dependence of the \ac{TSTDP} rule.

\begin{table}[H]
\centering
    \caption{Variables of the \ac{SBCM} rule.}
    \label{tab:sbcm}
    \begin{tabular}{@{}cc@{}}
    \toprule
    \textbf{Variable} & \textbf{Description} \cr
    \midrule
    $w_{ij}$ & Synaptic weight between pre- and post-synaptic neurons $i$ and $j$, respectively \cr
    $\kappa$ & Learning rate \cr
    $\alpha_j$ & Scaling factor (gain) associated with the neuron \cr
    $a_i$ / $a_j$ & Pre- and post-synaptic spike traces \cr
    $\theta(t)$ & Modification threshold: $\theta(t) = e^{-t / \tau} \theta(t-1) + (1 - e^{-t / \tau} a_j(t))$ \cr
    $\tau$& Time constant of modification threshold\cr
    \bottomrule
    \end{tabular}
\end{table}


\subsection{Yger and Harris (2013): \acf{MPDP}}
The \acf{MPDP} rule, also called the ``Convallis'' rule~\cite{Yger_Harris13} aims to approximate a fundamental computational principle of the neocortex and is derived from principles of unsupervised learning algorithms. The main assumption of the rule is that projections with non-Gaussian distributions are more likely to extract useful information from real-world patterns~\cite{Hyvarinen_Oja00}. Therefore, synaptic changes should tend to increase the skewness of a neuron’s sub-threshold membrane potential distribution. The rule is therefore derived from an objective function that measures how non-Gaussian the membrane potential distribution is, such that the post-synaptic neuron is often close to either its resting potential or spiking threshold (and not in between).

The resulting plasticity rule reinforces synapses that are active during post-synaptic depolarization and weakens those active during hyper-polarization. It is expressed in Eq.~\eqref{eq:mpdp-trace}, where changes are continuously made on an internal update trace $\Psi$, and are then applied on the synaptic weight $w$ as expressed in Eq.~\eqref{eq:mpdp-update}. The variables of the equations are explained in Tab.~\ref{tab:mpdp}.
The rule was used for unsupervised learning of speech data, where an additional mechanism was implemented to maintain a constant average firing rate.

\begin{equation}
\label{eq:mpdp-trace}
    \Psi(t) = \int_{-\infty}^{t} e^{-(t - \tau)/T} F'(V(\tau)) \sum_{i=1}^{N_s} K(\tau - t_i^s) d\tau
\end{equation}

\begin{equation}
\label{eq:mpdp-update}
    \frac{dw}{dt} = \left\{\begin{matrix}
    \Psi - \theta_{\mathrm{pot}} & if \: \theta_{\mathrm{pot}} < \Psi \\
    0 & if \: \theta_{\mathrm{dep}} < \Psi \leq \theta_{\mathrm{pot}} \\ 
    \Psi - \theta_{\mathrm{dep}} & if \: \Psi \leq \theta_{\mathrm{dep}} 
\end{matrix}\right.
\end{equation}

\begin{table}[H]
\centering
    \caption{Variables of the \ac{MPDP} rule.}
    \label{tab:mpdp}
    \begin{tabular}{@{}cc@{}}
    \toprule
    \textbf{Variable} & \textbf{Description} \cr
    \midrule
    $\Psi$ & Internal spike trace \cr
    $T$& Decay time constant\cr
    $F'(V(\tau))$ & Function of the post-synaptic membrane voltage \cr
    $V(\tau)$& Post-synaptic membrane voltage\cr
    $N_{s}$& Pre-synaptic spike indices\cr
    $\sum_{i=1}^{N_s} K(\tau - t_i^s)$ & Pre-synaptic spike trace - integrative \cr
    $K(\tau - t_i^s)$& Kernel for pre-synaptic spikes\cr
    $w$ & Synaptic weight \cr
    $\theta_{\mathrm{pot}}$ / $\theta_{\mathrm{dep}}$ & Thresholds for potentiation and depression\cr
    \bottomrule
    \end{tabular}
\end{table}


\subsection{Urbanczik and Senn (2014): \acf{DPSS}}
\citeasnoun{Urbanczik_Senn14} proposed a new learning model based on the \acf{DPSS}, which aims to implement a biologically plausible non-Hebbian learning rule. In their rule, they rely on the pre-synaptic spike trace, the post-synaptic spike event and the post-synaptic dendritic voltage of a multi-compartment neuron model.
Plasticity in dendritic synapses is realizing a predictive coding scheme that matches the dendritic potential to the somatic potential. This minimizes the error of dendritic prediction of somatic spiking activity of a conductance-based neuron model, that exhibits probabilistic spiking~\cite{Urbanczik_Senn14}.
The neuron membrane potential $U$ is influenced by both a scaled version of the dendritic compartment potential $V^{*}_{w}$ and the teaching inputs from excitatory or inhibitory proximal synapses $I_{U}^{\mathrm{som}}$.

In their proposed learning rule (see Eq.~\eqref{eq:dpss-PI}), the aim is to minimize the error between the predicted somatic spiking activity based on the dendritic potential $\phi(V_{w}^*(t))$ and the real somatic spiking activity represented by back-propagated spikes $S(t)$. The equation's variables are described in Tab.~\ref{tab:dpss}. The error $S(t)-\phi(V_{w}^*(t))$ is assigned to individual dendritic synapses based on their recent activation, similar to~\citeasnoun{Yger_Harris13} and~\citeasnoun{Albers_etal16}. 

\begin{equation}
        PI_{i}(t) = [S(t) - \phi(V_{w}^*(t))]h(V_{w}^*(t))PSP_i(t)
\label{eq:dpss-PI}
\end{equation}

Since the back-propagated spikes $S(t)$ are only $0$ or $1$, but the predicted rate $\phi(V_{w}^*)$ based on a sigmoidal function is never $0$ or $1$, $PI$ will never be $0$. In this case, there is never a zero weight change~\cite{Urbanczik_Senn14}. 
The plasticity induction variable $PI_{i}$ is continuously updated and used as an intermediate variable, 
before it is applied to induce a scaled 
persistent synaptic change, as expressed in Eq.~\eqref{eq:dpss-PI-lp}.

\begin{equation}
    \begin{split}
       \tau_{\Delta}\frac{d\Delta_{i}}{dt} &= PI_{i}(t) - \Delta_{i} \\
       \frac{dw_{i}}{dt} &= \eta\Delta_{i}
    \end{split}
\label{eq:dpss-PI-lp}
\end{equation}

\citeasnoun{Sacramento_etal18} showed later analytically that the \acf{DPSS} learning rule combined with similar dendritic predictive plasticity mechanisms approximate the error back-propagation algorithm, and demonstrated the capabilities of such a learning framework to solve regression and classification tasks.

\begin{table}[H]
\centering
    \caption{Variables of the \ac{DPSS} rule.}
    \label{tab:dpss}
    \begin{tabular}{@{}cc@{}}
    \toprule
    \textbf{Variable} & \textbf{Description} \cr
    \midrule
    $U$ & Somatic potential \cr
    $V^{*}_{w}$ & Scaled dendritic potential \cr
    $I_{U}^{\mathrm{som}}$ & Proximal input current \cr
    $\phi(.)$ & Sigmoid function \cr
    $S(t)$ & Back-propagated somatic spiking activity \cr
    $PI_i(t)$ & Plasticity induction variable \cr
    $h(.)$ & Positive weighting function \cr
    $PSP_i(t)$ & Pre-synaptic spike trace - integrative \cr
     & $PSP_i(t)=\sum_{s\in X_{i}^{\mathrm{dnd}}} \kappa(t-s)$\cr
    $\kappa(t-s)$ & Kernel for pre-synaptic spikes \cr
    $X^{\mathrm{dnd}}_i$ & Pre-synaptic spike train \cr
    $w_{i}$ & Synaptic strength of synapse $i$ \cr
    $\tau_{\Delta}$ & Plasticity induction variable time constant \cr
    $\eta$ & Learning rate \cr
    \bottomrule
    \end{tabular}
\end{table}


\subsection{Diehl and Cook (2015): \acf{RDSP}}
\citeasnoun{Diehl_Cook15} proposed the~\acf{RDSP} rule as a local credit assignment mechanism for unsupervised learning in self-organizing \acp{SNN}. The idea is to potentiate or depress the synapses for which the pre-synaptic neuron activity was high or low at the moment of a post-synaptic spike, respectively. 
The \ac{RDSP} learning rule relies solely on the pre-synaptic information and is triggered when a post-synaptic spike arrives. The weight update is shown in Eq.~\eqref{eq:rdsp}, whose variables are described in Tab.~\ref{tab:rdsp}.

\begin{equation}
\label{eq:rdsp}
    \Delta w = \eta (x_{\mathrm{pre}} - x_{\mathrm{tar}}) \: (w_{\mathrm{max}} - w)^{u}
\end{equation}

$u$ determines the weight dependence of the update for implementing a soft bound, while the target value of the pre-synaptic spike trace $x_{tar}$ is crucial in this learning rule because it acts as a threshold between depression and potentiation. If it is set to $0$, then only potentiation is observed. It is hence important to set it to a non-zero value to ensure that pre-synaptic neurons that rarely lead to the firing of the post-synaptic neuron will become more and more disconnected. More generally, the higher the value of $x_{\mathrm{tar}}$ value, the more depression occurs and the lower the synaptic weights will be~\cite{Diehl_Cook15}.

This rule was first proposed as a more biologically plausible version of a previously proposed rule for memristive implementations by~\citeasnoun{Querlioz_etal13}. 
The main difference between the two models is that the \ac{RDSP} rule uses an exponential time dependence for the weight change which is more biologically plausible~\cite{Abbott_Song99} than a time-independent weight change. This can also be more useful for pattern recognition depending on the temporal dynamics of the learning task.

\begin{table}[H]
\centering
    \caption{Variables of the \ac{RDSP} rule.}
    \label{tab:rdsp}
    \begin{tabular}{@{}cc@{}}
    \toprule
    \textbf{Variable} & \textbf{Description} \cr
    \midrule
    $w$& Synaptic weight\cr
    $\eta$& Learning rate\cr
    $x_{\mathrm{pre}}$& Pre-synaptic spike trace - integrative\cr
    $x_{\mathrm{tar}}$& Target value of the pre-synaptic spike trace\cr
    $w_{\mathrm{max}}$& Maximum weight\cr
    $u$& Weight dependence - soft bound\cr
    \bottomrule
    \end{tabular}
\end{table}


\subsection{Albers et al. (2016): \acf{HMPDP}}
The \acf{HMPDP} learning rule proposed by~\citeasnoun{Albers_etal16} is derived from an objective function similar to that of the \ac{MPDP} rule but with opposite sign, as it aims to balance the membrane potential of the post-synaptic neuron between two fixed thresholds; the resting potential and the spiking threshold of the neuron. Hence, the \ac{MPDP} and the \ac{HMPDP} implement a Hebbian or homeostatic mechanism, respectively. In addition, the \ac{HMPDP} differs from the other described models by inducing plasticity only to inhibitory synapses.

\citeasnoun{Albers_etal16} use a conductance based neuron and synapse model, similar to the \acs{CMPDP} and the \acs{DPSS} rules. The continuous weight updates of the \ac{HMPDP} rule depend on the instantaneous membrane potential $V(t)$ and the pre-synaptic spike trace $\sum_{k} \epsilon(t-t_{i}^{k})$ as expressed in Eq.~\eqref{eq:hmpdp} whose variables are described in Tab.~\ref{tab:hmpdp}.

\begin{equation}
\label{eq:hmpdp}
    w_i = \eta (-\gamma[V(t) - \vartheta_D]_+ + [\vartheta_P - V(t)]_+)\sum_{k} \epsilon(t-t_{i}^{k})
\end{equation}

The authors claim that their model is able to learn precise spike times by keeping a homeostatic membrane potential between two thresholds. This definition differs from the homeostatic spike rate definition of the \acs{CMPDP} rule by~\citeasnoun{Sheik_etal16}.

\begin{table}[H]
\centering
    \caption{Variables of the \ac{HMPDP} rule.}
    \label{tab:hmpdp}
    \begin{tabular}{@{}cc@{}}
    \toprule
    \textbf{Variable} & \textbf{Description} \cr
    \midrule
    $w_i$& Synaptic weight \cr
    $\eta$& Learning rate \cr
    $\gamma$& Scaling factor for LTD/LTP\cr
    $[.]_{+}$& Rectifying bracket $[x]_+ = x$ if $x>0$, $[x]_+ =0$ otherwise\cr
    $V(t)$& Instantaneous membrane potential\cr
    $\vartheta_P/\vartheta_D$& Thresholds for plasticity induction\cr
    $\sum_{k} \epsilon(t-t_{i}^{k})$& Pre-synaptic spike trace - integrative\cr
    $t_{i}^{k}$& Time of the k-th spike at the i-th synapse\cr
    $\epsilon(s)$& Kernel for pre-synaptic spikes \cr
    \bottomrule
    \end{tabular}
\end{table}


\subsection{Sheik et al. (2016): \acf{CMPDP}}
The \acf{CMPDP} learning rule~\cite{Sheik_etal16} was proposed with the explicit intention to have a local spike-timing based rule that would be sensitive to the order of spikes arriving at different synapses and that could be ported onto neuromorphic hardware. 

Similarly to the \ac{DPSS} rule, the \ac{CMPDP} rule uses a conductance-based neuron model. However, instead of relying on mean rates, it relies on the exact timing of the spikes. Furthermore,  as for the \ac{HMPDP} rule, \citeasnoun{Sheik_etal16} propose to add a homeostatic element to the rule that targets a desired output firing rate. 
This learning rule is very hardware efficient because it depends only on the pre-synaptic spike time and not on the post-synaptic one. 
The equation that governs its behavior is Eq.~\eqref{eq:cmpdp-update}.
The weight update, triggered by the pre-synaptic spike, depends on a membrane voltage component (see Eq.~\eqref{eq:cmpdp-update-voltage}) and on a homeostatic one (see Eq.~\eqref{eq:cmpdp-update-homo}). 
All equation variables are described in Tab.~\ref{tab:cmpdp}.

\begin{equation}
\Delta W= \Delta W_{v} + \Delta W_{h}
\label{eq:cmpdp-update}
\end{equation}

\begin{equation}
\Delta W_{v}=[\delta(V_{m}(t+1)>V_{\mathrm{lth}}) \eta_{+} -\delta(V_{m}(t+1)<V_{\mathrm{lth}}) \eta_{-}] S(t-t_{\mathrm{pre}})
\label{eq:cmpdp-update-voltage}
\end{equation}

\begin{equation}
\Delta W_{h}=\eta_{h}(Ca_{t}-Ca) S(t-t_{\mathrm{pre}})
\label{eq:cmpdp-update-homo}
\end{equation}

The post-synaptic membrane voltage dependent weight update shown in Eq.~\eqref{eq:cmpdp-update-voltage} depends on the values of the membrane voltage $V_{m}$ and an externally set threshold $V_{\mathrm{lth}}$, which determines the switch between \ac{LTP} and \ac{LTD}. 
The homeostatic weight update in Eq.~\eqref{eq:cmpdp-update-homo} is proportional to the difference in post-synaptic activity represented by the post-synaptic spike trace $Ca$ and an externally set threshold $Ca_{t}$. 

The authors show that this learning rule, using the spike timing together with conductance based neurons, is able to learn spatio-temporal patterns in noisy data and differentiate between inputs that have the same 1st-moment statistics but different higher moment ones. 
Although they gear the rule toward neuromorphic hardware implementations, they do not propose circuits for the learning rule.

\begin{table}[H]
\centering
    \caption{Variables of the \ac{CMPDP} rule.}
    \label{tab:cmpdp}
    \begin{tabular}{@{}cc@{}}
    \toprule
    \textbf{Variable} & \textbf{Description} \cr
    \midrule
    $W$& Synaptic weight\cr
    $\Delta W_{v}$& Voltage-based weight update\cr
    $\Delta W_{h}$& Homeostatic weight update\cr
    $\delta$& Boolean variable\cr
    $V_{m}$& Membrane potential \cr
    $V_{\mathrm{lth}}$& Threshold on membrane potential \cr
    $\eta_{+}$ / $\eta_{-}$ / $\eta_{h}$ & Magnitude of LTP/LTD/Homeostasis\cr
    $S(t-t_{\mathrm{pre}})$& Pre-synaptic spike trains\cr
    $t_{\mathrm{pre}}$& Pre-synaptic spike time\cr
    $Ca$& Post-synaptic spike trace (Calcium) - integrative\cr
    $Ca_{t}$& Calcium target concentration trace\cr
    \bottomrule
    \end{tabular}
\end{table}


\subsection{Payeur et al. (2021): \acf{BDSP}}
The \acf{BDSP} learning rule~\cite{Payeur_etal21} has been proposed to enable online, local, spike-based solutions to the credit assignment problem in hierarchical networks~\cite{Zenke_Neftci21}, i.e.\ how can neurons high up in a hierarchy signal to other neurons, sometimes multiple synapses apart, whether to engage in \ac{LTP} or \ac{LTD} to improve behavior. 
The \ac{BDSP} learning rule is formulated in Eq.~\eqref{eq:bdsp} whose variables are described in Tab.~\ref{tab:bdsp}.

\begin{equation}
\label{eq:bdsp}
    \frac{dw_{ij}}{dt} = \eta [B_i(t) - \overline{P}_i(t) E_i(t)] \widetilde{E}_j(t)
\end{equation}

where an event $E_i(t)$ is said to occur either at the time of an isolated spike or at the time of the first spike in a burst, whereas a burst $B_i(t)$ is defined as any occurrence of at least two spikes (at the second spike) with an inter-spike interval less than a pre-defined threshold. Any additional spike within the time threshold belongs to the same burst. Hence, \ac{LTP} and \ac{LTD} are triggered by a burst and an event, respectively. Since a burst is always preceded by an event, every potentiation is preceded by a depression. However, the potentiation through the burst is larger than the previous depression, which results in an overall potentiation.

The moving average $\overline{P_i}(t)$ regulates the relative strength of burst-triggered potentiation and event-triggered depression. It has been established that such a mechanism exists in biological neurons~\cite{Maki-Marttunen_etal20}. It is formulated as a ratio between averaged post-synaptic burst and event traces.
The authors show that manipulating the moving average $\overline{P_i}(t)$ (i.e.\ the probability that an event becomes a burst) controls the occurrence of \ac{LTP} and \ac{LTD}, while changing the pre- and post-synaptic event rates simply modifies the rate of change of the weight while keeping the same transition point between \ac{LTP} and \ac{LTD}. Hence, the \ac{BDSP} rule paired with the control of bursting provided by apical dendrites enables a form of top-down steering of synaptic plasticity in an online, local and spike-based manner.

Moreover, the authors show that this dendrite-dependent bursting combined with short-term plasticity supports multiplexing of feed-forward and feedback signals, which means that the feedback signals can steer plasticity without affecting the communication of bottom-up signals. Taken together, these observations show that combining the \ac{BDSP} rule with short-term plasticity and apical dendrites can provide a local approximation of the credit assignment problem. In fact, the learning rule has been shown to implement an approximation of gradient descent for hierarchical circuits and achieve good performance on standard machine learning benchmarks.

\begin{table}[H]
\centering
    \caption{Variables of the \ac{BDSP} rule.}
    \label{tab:bdsp}
    \begin{tabular}{@{}cc@{}}
    \toprule
    \textbf{Variable} & \textbf{Description} \cr
    \midrule
    $w_{ij}$ & Synaptic weight between pre- and post-synaptic neurons $j$ and $i$ \cr
    $\eta$ & Learning rate \cr
    $B_i(t)$ & Post-synaptic bursts \cr
    $\overline{P}_i(t)$ & Exponential moving average of the proportion of post-synaptic bursts \cr
    $E_i(t)$ & Post-synaptic events \cr
    $\widetilde{E}_j(t)$ & Pre-synaptic spike trace \cr
    \bottomrule
    \end{tabular}
\end{table}


\subsection{Models common variables}
Tables \ref{tab:models-variablesI} and \ref{tab:models-variablesII} show the major common variables between the different models. This allows an easy comparison of the formalism of each rule.

\begin{center}
    \begin{table}[H]

    \caption{Variables in common between rules Part I}
    \label{tab:models-variablesI}
        \resizebox{\linewidth}{!}{%
            \begin{tabular}{>{\centering}m{0.15\linewidth}>{\centering}m{0.15\linewidth}>{\centering}m{0.15\linewidth}>{\centering}m{0.15\linewidth}>{\centering}m{0.15\linewidth}>{\centering}m{0.15\linewidth}>{\centering}m{0.15\linewidth}}
            \toprule
            \textbf{Variables}&\textbf{\acs{STDP}}&\textbf{\acs{TSTDP}}&\textbf{\acs{SDSP}}&\textbf{\acs{VSTDP}}&\textbf{\acs{CSTDP}}&\textbf{\acs{SBCM}}\cr
            \midrule
            Synaptic weight &w &w & X & $w$ & $\rho$ & $w_{ij}$\cr
            \midrule
            Weight bounds& & & $X_{max}$ & $w_{max}$&  $0$ / $1$\cr
            \midrule
            Traces& &$o_{1}$ / $o_{2}$ / $r_{1}$ / $r_{2}$ & $C(t)$ & $\overline{u}_{-}(t)$ / $\overline{u}_{+}(t)$ / $\overline{x}(t)$ & $c(t)$ & $a_i$ / $a_j$\cr
            \midrule
            Time constants & $\tau_{+}$ / $\tau_{-}$ & & & & $\tau$ & $\tau$ \cr
            \midrule
            Membrane potential& & &$V(t)$ & $u(t)$ &\cr
            \midrule
            Thresholds & & & $\theta_{V}$ / $\theta_{up}^{\mathrm{l}}$ / $\theta_{up}^{\mathrm{h}}$ / $\theta_{down}^{\mathrm{l}}$ / $\theta_{down}^{\mathrm{h}}$ / $\theta_{X}$ & $\theta_{-}$ / $\theta_{+}$ & $\rho_{\star}$ / $\theta_p$ / $\theta_d$ &$\theta$\cr
            \midrule
            Amplitudes & $\mathrm{A}_{+}$ / $\mathrm{A}_{-}$ & $\mathrm{A}_{2+}$ / $\mathrm{A}_{2-}$ / $\mathrm{A}_{3+}$ / $\mathrm{A}_{3-}$ & $a$ / $b$ / $\alpha$ / $\beta$ & $A_{LTP}$ / $A_{LTD}$ & $C_{pre}$ / $C_{post}$ / $\gamma_{p}$ / $\gamma_{d}$ & $\kappa$ / $\alpha_j$ \cr
            \bottomrule
            \end{tabular}
        }

    \end{table}
\end{center}

\begin{center}
    \begin{table}[H]

    \caption{Variables in common between rules Part II }
    \label{tab:models-variablesII}
        \resizebox{\linewidth}{!}{%
            \begin{tabular}{>{\centering}m{0.155\linewidth}>{\centering}m{0.125\linewidth}>{\centering}m{0.125\linewidth}>{\centering}m{0.125\linewidth}>{\centering}m{0.15\linewidth}>{\centering}m{0.15\linewidth}>{\centering}m{0.125\linewidth}}
            \textbf{Variables}&\textbf{\acs{MPDP}}&\textbf{\acs{DPSS}}&\textbf{\acs{RDSP}}&\textbf{\acs{HMPDP}}&\textbf{\acs{CMPDP}}&\textbf{\acs{BDSP}}\cr
            \midrule
            Synaptic weight & $w$ & $w_{i}$ & $w$ & $w_i$ & $W$ & $w_{ij}$ \cr
            \midrule
            Weight bounds & & & $w_{max}$ &  & &  \cr
            \midrule
            Traces & $\Psi$ / $ K(\tau - t_i^s)$ & $PSP_i(t)$ / $\Delta_{i}$ & $x_{pre}$ & $\sum_{k} \epsilon(t-t_{i}^{k})$ & $Ca$ / $Ca_{t}$ & $\widetilde{E_j}(t)$ \cr
            \midrule
            Time constants & T & $\tau_{\Delta}$ & & & &  \cr
            \midrule
            Membrane potential & $V(\tau)$ & $U$ & & $V(t)$ & $V_{m}$ & \cr
            \midrule
            Thresholds & $\theta_{dep}$ / $\theta_{pot}$  & 
            & $x_{tar}$ & $\vartheta_P$ / $\vartheta_D$ & $V_{lth}$ & \cr
            \midrule
            Amplitudes & & $\eta$ 
            &$\eta$ / $u$ & $\eta$ / $\gamma$ & $\eta_{+}$ / $\eta_{-}$ / $\eta_{h}$ & $\eta$ / $\overline{P_i}(t)$ \cr
            \bottomrule
            \end{tabular}
        }

    \end{table}
\end{center}


\section{\ac{CMOS} implementations of synaptic plasticity}
Our comparison of plasticity models has highlighted many common functional primitives that are shared among the rules. These primitives can be grouped according to their function into the following blocks: low-pass filters, eligibility traces, and weight updates. 
These blocks can be readily implemented in \ac{CMOS} technology, and they can be combined to implement different learning circuits.
An overview of the proposed \ac{CMOS} learning circuits that implement some of the models discussed is shown in Table~\ref{tab:circuits}.
To better link the \ac{CMOS} implementations with the models presented, we named all the current and voltage variables of our circuits to match those in the model equations.


\subsection{\ac{CMOS} building blocks}
The basic building blocks found required for building neuromorphic learning circuits can be grouped in four different families. 

\begin{description}
    \item[Eligibility trace blocks] These are implemented using either a current-mode integrator circuit, such as the \acf{DPI}, or other non-linear circuits that produce slowly decaying signals. 
    Input spikes can either increase the trace amplitude, decrease it, or completely reset it. The rate at which the trace decays back to its resting state can be typically modulated with externally controllable parameters. 
    Circuit blocks implementing eligibility traces are highlighted in green in the schematics.

    \item [Comparator blocks] They are typically implemented using \acf{WTA} current mode circuits, or voltage mode transconductance or \acp{OPAMP}. The comparator block changes its output based on which input is greater. Circuit blocks implementing comparators are highlighted in yellow in the schematics. 
    
    \item [Weight update blocks] They typically comprise a capacitor that stores a voltage related to the amplitude of the weight. Charging and discharging pathways connected to the capacitor enable potentiation and depression of the weight depending on the status of other signals.
    These blocks are is similar to the eligibility trace ones, except for the fact that they can produce both positive and negative changes.  
    Circuit blocks implementing weight updates are highlighted in purple in the schematics. 
    
    \item[Bistability blocks] These are typically implemented using a \ac{TA} connected in feedback operation which compares the weight voltage to a reference voltage. 
    Depending on the value of the weight voltage the bistability circuit will push the weight to the closest stable state.
    In its simplest form they have one single reference voltage, but they could be expanded to produce multiple stable states. 
    Circuit blocks implementing bistability are highlighted in red in the schematics.
\end{description}

\begin{center}
    \begin{table}[H]
        \caption{Neuromorphic circuits for spike-based local synaptic plasticity models}
        \label{tab:circuits}
        \resizebox{\textwidth}{!}{%
        \begin{tabular}{>{\hspace{0pt}}m{0.1\linewidth}>{\centering\hspace{0pt}}m{0.3\linewidth}>{\centering\hspace{0pt}}m{0.37\linewidth}>{\centering\hspace{0pt}}m{0.23\linewidth}} 
        \hline
        \textbf{Rule} & \textbf{Paper} & \textbf{Difference with the model} & \textbf{Implementation} \cr
        \hline
        \multirow{25}{*}{\acs{STDP}} &~\cite{Bofill-i-Petit_etal01}$^1$ & / & \SI{0.6}{\um} Fabricated \cr
        \cline{2-4}
         &~\cite{Indiveri02b} & All-to-all spike interaction + bistable weights & \SI{1.5}{\um} Fabricated \cr
         \cline{2-4}
         &~\cite{Bofill-i-Petit_Murray04} & / & \SI{0.6}{\um} Fabricated \cr
         \cline{2-4}
         &~\cite{Cameron_etal05} & Anti-STDP + Non-exponential spike trace & \SI{0.35}{\um} Fabricated \cr
         \cline{2-4}
         &~\cite{Indiveri_etal06} & Bistable weights & \SI{1.6}{\um} Fabricated \cr
         \cline{2-4}
         &~\cite{Arthur_Boahen06}$^2$ & All-to-all interaction + binary weights & \SI{0.25}{\um} Fabricated \cr
         \cline{2-4}
         &~\cite{Koickal_etal07} & Soft bounds & \SI{0.6}{\um} Fabricated \cr
         \cline{2-4}
         &~\cite{Liu_Mockel08} & All-to-all spike interaction + asymmetric bounds (soft lower bound + hard upper bound) & \SI{0.35}{\um} Fabricated \cr
         \cline{2-4}
         &~\cite{Tanaka_etal09} & / & \SI{0.25}{\um} Fabricated \cr
         \cline{2-4}
         &~\cite{Bamford_etal12} & All-to-all spike interaction & \SI{0.35}{\um} Fabricated \cr
         \cline{2-4}
         &~\cite{Gopalakrishnan_Basu14} & All-to-all spike interaction + asymmetric bounds (soft lower bound + hard upper bound) & \SI{0.35}{\um} Fabricated \cr
         \cline{2-4}
         &~\cite{Mastella_etal20} & / & \SI{0.15}{\um} Simulated \cr
        \hline
        \multirow{3}{*}{\acs{TSTDP}} &~\cite{Mayr_etal10} & / & Simulated \cr
        \cline{2-4}
         &~\cite{Azghadi_etal13} & / & \SI{0.35}{\um} Simulated \cr
        \cline{2-4}
         &~\cite{Gopalakrishnan_Basu17} & / & \SI{0.35}{\um} Fabricated \cr
        \hline
        \multirow{6}{*}{\acs{SDSP}} &~\cite{Fusi_etal00} & No post-synaptic spike trace + no stop-learning mechanism & \SI{1.2}{\um} Fabricated \cr
        \cline{2-4}
         &~\cite{Chicca_Fusi01} & No post-synaptic spike trace + no stop-learning mechanism & \SI{0.6}{\um} Fabricated \cr
        \cline{2-4}
         &~\cite{Chicca_etal03} & No post-synaptic spike trace + no stop-learning mechanism & \SI{0.6}{\um} Fabricated \cr
        \cline{2-4}
         &~\cite{Giulioni_etal08} & Analog weights & \SI{0.35}{\um} Fabricated \cr
        \cline{2-4}
         &~\cite{Mitra_etal09} & Analog weights & \SI{0.35}{\um} Fabricated \cr
        \cline{2-4}
         &~\cite{Chicca_etal14b} & Analog weights & \SI{0.35}{\um} Fabricated \cr
        \hline
        \multirow{1}{*}{\acs{CSTDP}} &~\cite{Maldonado_etal16} & Hard bounds & \SI{0.18}{\um} Fabricated \cr
        \hline
        \multirow{4}{*}{\acs{RDSP}} &~\cite{Hafliger_etal97} & Nearest spike interaction + reset of pre-synaptic spike trace at post-spike + very small soft bounds & \SI{2}{\um} Fabricated \cr
        \cline{2-4}
         &~\cite{Ramakrishnan_etal11} & Nearest spike interaction + asymmetric bounds (soft lower bound + hard upper bound) & \SI{0.35}{\um} Fabricated \cr
        \hline
        \end{tabular}%
        }
        \footnotesize{$^1$ Potentiation and depression triggers done with digital logic gates. \newline
        $^2$ Weight storage in digital SRAM.}
    \end{table}
\end{center}


\subsection{\acf{STDP}}

\begin{center}
    \begin{figure}[H]
        \includegraphics[width=0.7\textwidth]{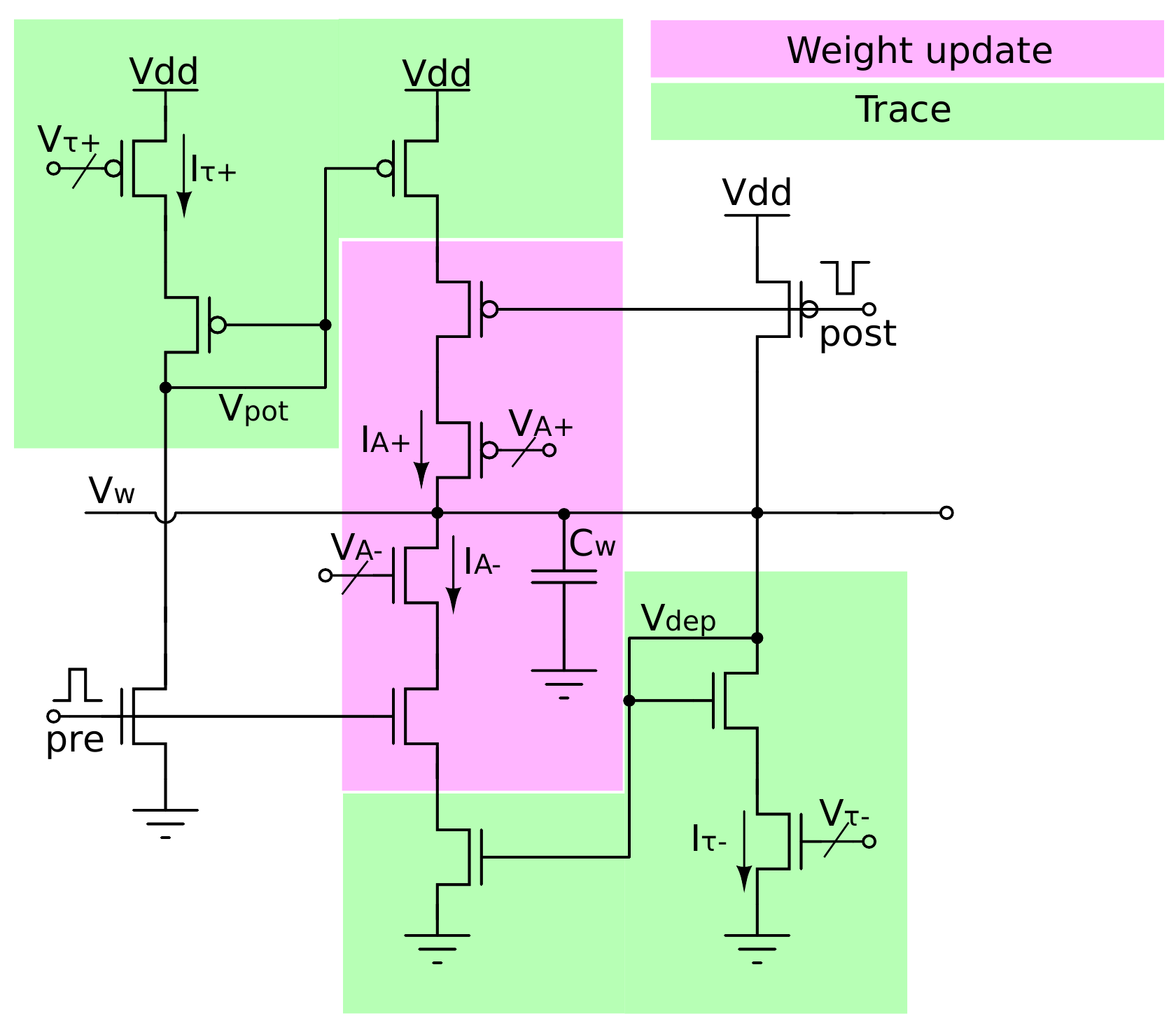}
        \centering
        \caption{\acs{STDP} circuit with highlighted the \ac{CMOS} building blocks used: Eligibility traces (in green) and Weight updates (in violet). The voltage and current variables reflect the model equation. Adapted from:~\protect\citeasnoun{Indiveri_etal06}.}
        \label{fig:stdp}
    \end{figure} 
\end{center}

Following the formalization of the \ac{STDP} model in 2000 (see Eq.~\eqref{eq:stdp}), many \ac{CMOS} implementations have been proposed. Most implement the model as explained in Section above~\cite{Bofill-i-Petit_etal01,Indiveri03,Bofill-i-Petit_Murray04,Arthur_Boahen06,Bamford_etal12} however, some exploit the physics of single transistors to propose a floating gate implementation~\cite{Liu_Mockel08,Gopalakrishnan_Basu14}.

\citeasnoun{Indiveri_etal06} presented the implementation in Fig.~\ref{fig:stdp}. 
This circuit increases or decreases the analog voltage $V_{w}$ across the capacitor $C_{w}$ depending on the relative timing of the pulses $pre$ and $post$. Upon arrival of a pre-synaptic pulse $pre$, a potentiating waveform $V_{pot}$ is generated within the pMOS-based trace block (see Fig.~\ref{fig:stdp}). $V_{pot}$ has a sharp onset and decays linearly with an adjustable slope set by $V_{\tau +}$. $V_{pot}$ serves to keep track of the most recent pre-synaptic spike.  Analogously, when a post-synaptic spike ($post$) occurs, $V_{dep}$ and $V_{\tau -}$ create a trace of post-synaptic activity. By ensuring that $V_{pot}$ and $V_{dep}$ remain below the threshold of the transistors they are connected to and the exponential current-voltage relation in the sub-threshold regime, the exponential relationship to the spike time difference $\Delta t$ of the model is achieved. While $V_{A+}$ and $V_{A-}$ set the upper-bounds of the amount of current that can be injected or removed from $C_{w}$, the decaying traces $V_{pot}$ and $V_{dep}$ determine the value of $I_{A+}$ or $I_{A-}$ and ultimately the weight increase or decrease on the capacitor $C_{w}$ within the weight update block (see Fig.~\ref{fig:stdp}).


\subsection{\acf{TSTDP}}

\begin{center}
    \begin{figure}[H]
        \includegraphics[width=\textwidth, angle = 0 ]{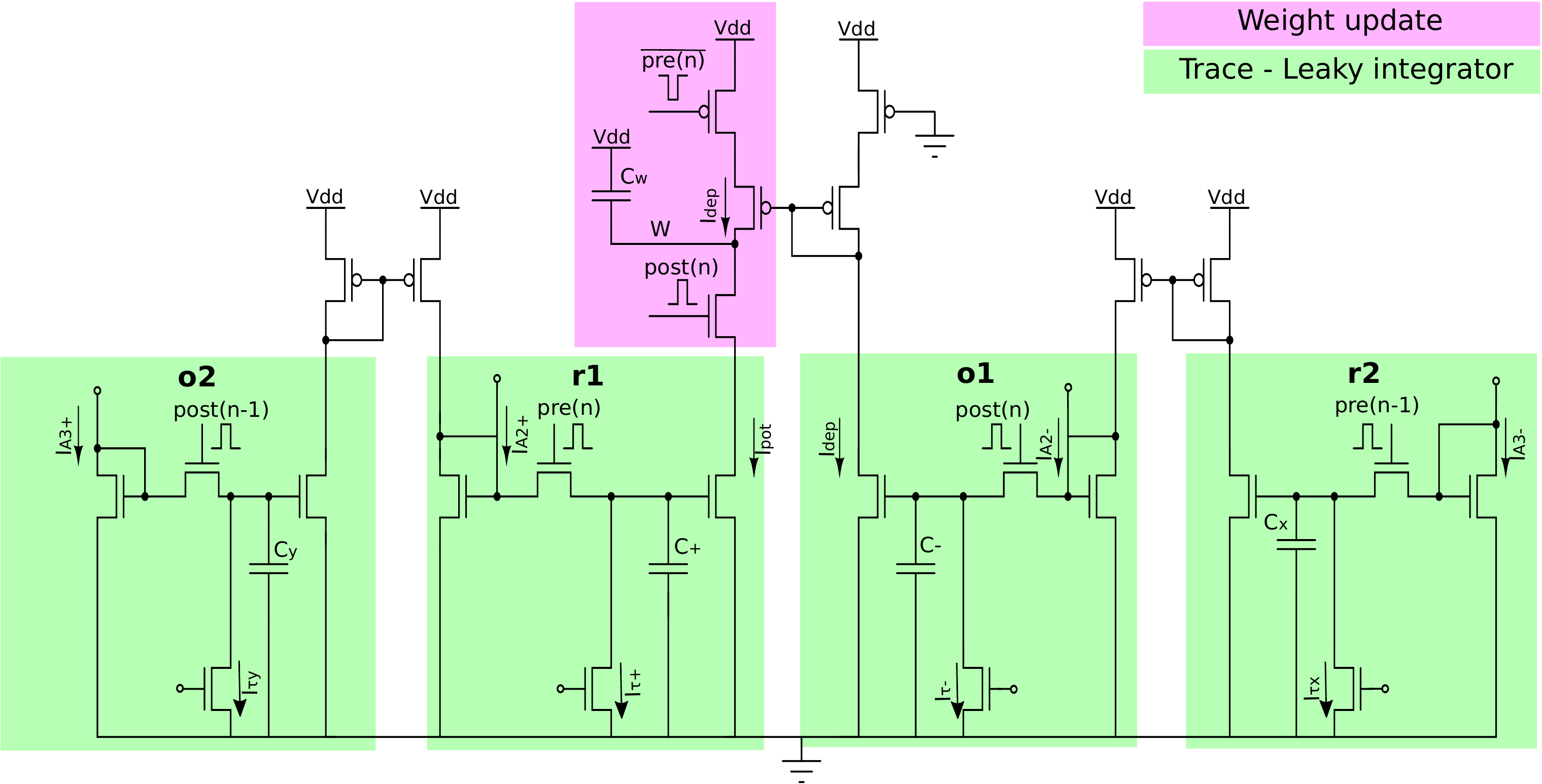}
        \centering
        \caption{\acs{TSTDP} circuit with highlighted the \ac{CMOS} building blocks used: Eligibility traces with leaky integrators (in green) and weight updates (in violet). The voltage and current variables reflect the model equation. The $r$ and $o$ detectors of the model are also reported in this circuit figure. Adapted from:~\protect\citeasnoun{Azghadi_etal13}.}
        \label{fig:tstdp}
    \end{figure}
\end{center}

Similarly, as for the pair-based \ac{STDP}, there are many implementations of the \ac{TSTDP} rule. While some are successful in implementing the equations in the model~\cite{Mayr_etal10,Meng_etal11,Rachmuth_etal11,Azghadi_etal13}, others exploit the properties of floating gates~\cite{Gopalakrishnan_Basu17}.
 
Specifically,~\citeasnoun{Mayr_etal10} as well as~\citeasnoun{Rachmuth_etal11} and~\citeasnoun{Meng_etal11} implement learning rules that model the conventional pair-based \ac{STDP} together with the \ac{BCM} rule. \citeasnoun{Azghadi_etal13} is the first, to our knowledge, to not only model the function but also model the equations presented in~\citeasnoun{Pfister_etal06} (see Eq.~\eqref{eq:tstdp}).
Figure~\ref{fig:tstdp} shows the circuit proposed by Azghadi in 2013 to model the \ac{TSTDP} rule. It faithfully implements the equations by having independent circuits and biases, for the model parameters $A_{2}^{-}$, $A_{2}^{+}$, $A_{3}^{-}$, and $A_{3}^{+}$. These parameters correspond to spike-pairs or spike-triplets: post-pre, pre-post, pre-post-pre, and post-pre-post, respectively. 

In this implementation, the voltage across the capacitor $C_{w}$ determines the weight of the specific synapse. Here, a high potential at the node $W$ is caused by a highly discharged capacitor indicating a low synaptic weight, which results in a depressed synapse. In the same way, a low potential at this node is caused by a more strongly charged capacitor and resembles a strong synaptic weight and in turn a potentiated synapse. The capacitor is charged and discharged by the two currents $I_{pot}$ and $I_{dep}$ respectively. These two currents are gated by the most recent pre- and post-synaptic spikes through the transistors controlled by $\overline{pre(n)}$ and $post(n)$ within the weight update block (see Fig.~\ref{fig:tstdp})

The amplitude of the depression current $I_{dep}$ and the potentiation current $I_{pot}$ is given by the recent spiking activity of the pre- and post-synaptic neurons. On the arrival of a pre-synaptic spike, the capacitors $C_{+}$ and $C_{x}$ (in the trace - leaky integrator blocks r1 and r2 in Fig.~\ref{fig:tstdp}) are charged by the currents $I_{A2+}$ and $I_{A3-}$. Analogously, the capacitors $C_{-}$ and $C_{y}$ (in the trace - leaky integrator blocks o1 and o2 in Fig.~\ref{fig:tstdp}) are charged at the arrival of a post-synaptic spike by the currents $I_{A2-}$ and $I_{A3+}$. Here, both currents $I_{A2+}$ and $I_{A2-}$ depend on an externally set constant input current plus the currents generated by the o2 and r2 blocks, respectively. These additional blocks o2 and r2 activated by previous spiking activity realize the triplet-sensitive behavior of the rule.
All capacitors within the``Trace - leaky integrator'' blocks ($C_{+}$, $C_{-}$, $C_{x}$, $C_{y}$) constantly discharge with individual rates given by $I_{\tau+}$, $I_{\tau-}$, $I_{\tau x}$, $I_{\tau y}$, respectively.


\subsection{\acf{SDSP}}

\begin{center}
    \begin{figure}[H]
        \includegraphics[width=\textwidth, angle = 0]{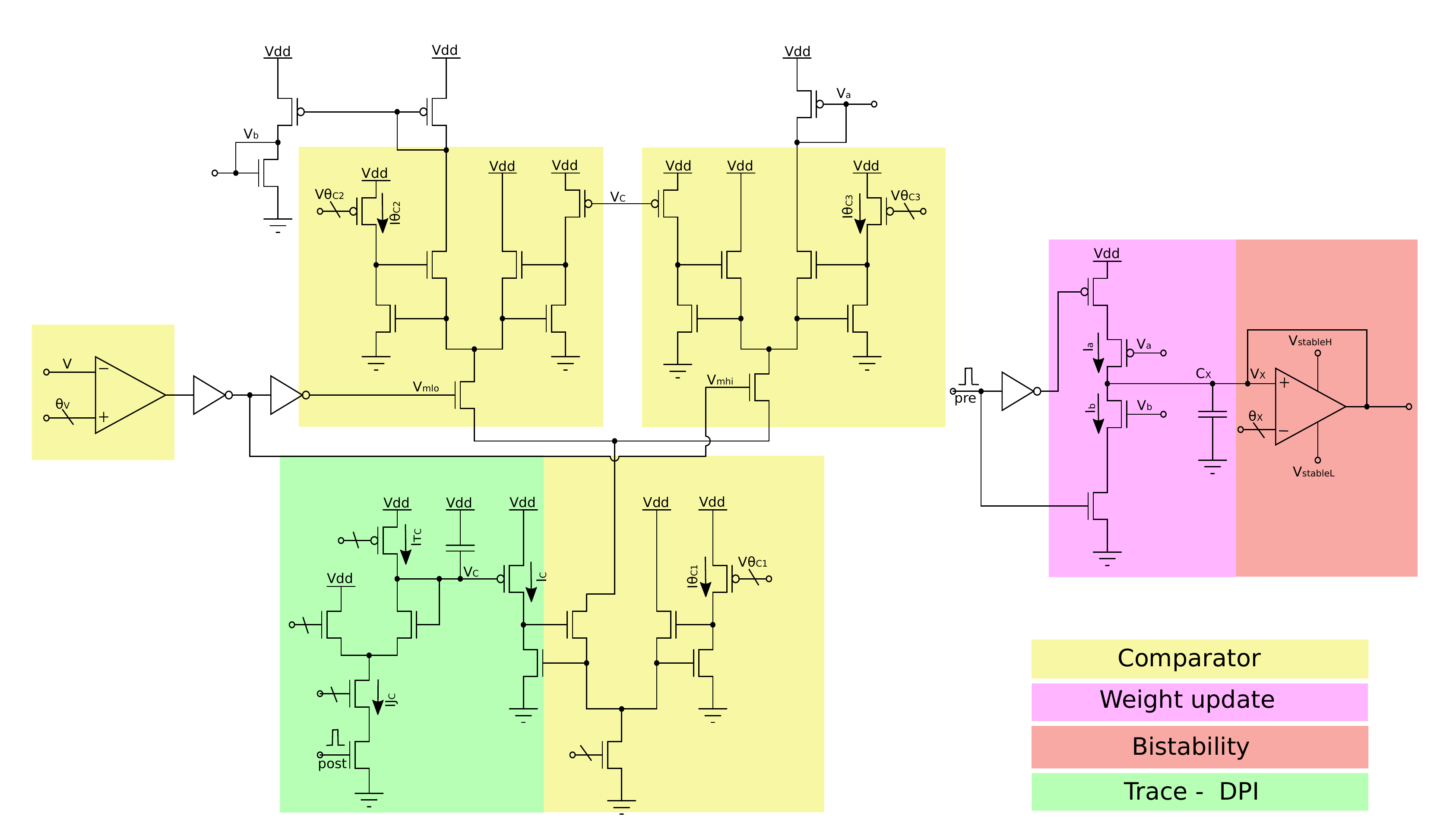}
        \centering
        \caption{\acs{SDSP} circuit with highlighted the \ac{CMOS} building blocks used: Eligibility traces with a \acs{DPI} (in green), weight updates (in violet), bistability (in red) and comparators with \acs{WTA} (in yellow). The voltage and current variables reflect the model equation. Adapted from:~\protect\citeasnoun{Chicca_etal14b}.}
        \label{fig:sdsp}
    \end{figure}
\end{center}

A sequence of theoretical works on spike based learning rules designed in the theoretical framework of attractor neural network and mean field theory preceded the \acf{SDSP} formalization by~\citeasnoun{Brader_etal07}. Several hardware implementations by~\citeasnoun{Fusi_etal00},~\citeasnoun{Dante_etal01} and~\citeasnoun{Chicca_etal03} accompanied this theoretical work. After formalization by~\citeasnoun{Brader_etal07} many implementations of the \acf{SDSP} rule were proposed following the desire to build smarter, larger, and more autonomous networks.
The implementations by~\citeasnoun{Chicca_etal03},~\citeasnoun{Mitra_etal09},~\citeasnoun{Giulioni_etal08} and~\citeasnoun{Chicca_etal14b} share similar building blocks: trace generators, comparators, blocks implementing the weight update and bistability mechanism. Here, we present the most complete design by~\citeasnoun{Chicca_etal14b}, shown in Fig.~\ref{fig:sdsp}, which replicates more closely the model equations (see Eq.~\eqref{eq:sdsp}).

At each pre-synaptic spike $pre$, the weight update block (see Fig.~\ref{fig:sdsp}) charges or discharges the capacitor $C_{x}$ altering the voltage $V_{x}$, depending on the values of $V_{a}$ and $V_{b}$. Here, $V_{x}$ represents the synaptic weight. If $I_{a} > I_{b}$, $V_{x}$ increases, while in the opposite case $V_{x}$ decreases. Moreover, over long time scales, in the absence of pre-synaptic spikes, $V_{x}$ is slowly driven toward the bistable states $V_{stableH}$ or $V_{stableL}$ depending on whether $V_{x}$ is higher or lower than $\theta_{x}$ respectively (see bistability block in Fig.~\ref{fig:sdsp}).

The $V_{a}$ and $V_{b}$ signals are continuously computed in the learning block, which compares the membrane potential of the neuron ($V$) to the threshold $\theta_{V}$ and evaluates in which region the Calcium concentration $V_{c}$ lies. The neuron's membrane potential is compared to the threshold $\theta_{V}$ by a transconductance amplifier. If $V > \theta_{V}$, $V_{mhi}$ is high and $V_{mlo}$ is low, while if $V < \theta_{V}$, $V_{mhi}$ is low and $V_{mlo}$ is high. At the same time, the post-synaptic neuron spikes ($post$) are integrated by a \ac{DPI} to produce the Calcium concentration $V_{c}$ (see trace - \ac{DPI} block in Fig.~\ref{fig:sdsp}), which is then compared with three Calcium thresholds by three \ac{WTA} circuits (see comparator circuits in Fig.~\ref{fig:sdsp}). In the lower comparator, $I_{c}$ is compared to $I_{\theta{C1}}$ and if $I_{c} < I_{\theta{C1}}$ no learning conditions of the \ac{SDSP} rule is satisfied and there is no weight update. Assuming that $I_{c} > I_{\theta{C1}}$, the two upper comparators set the signals $V_{a}$ and $V_{b}$. If $V_{mlo}$ is high and $I_{c} < I_{\theta{C2}}$, $V_{b}$ is increasing, setting the strength of the nMOS-based pull-down branch in the weight update block. If $V_{mhi}$ is high and $I_{C} < I_{\theta{C3}}$, $V_{a}$ is decreasing, setting the strength of the pMOS-based pull-up branch of the weight update block. These two branches in the weight update block are activated by the $pre$ input spike.


\subsection{\acf{CSTDP}}

\begin{center}
    \begin{figure}[H]
        \includegraphics[width=0.8\textwidth,  angle =0]{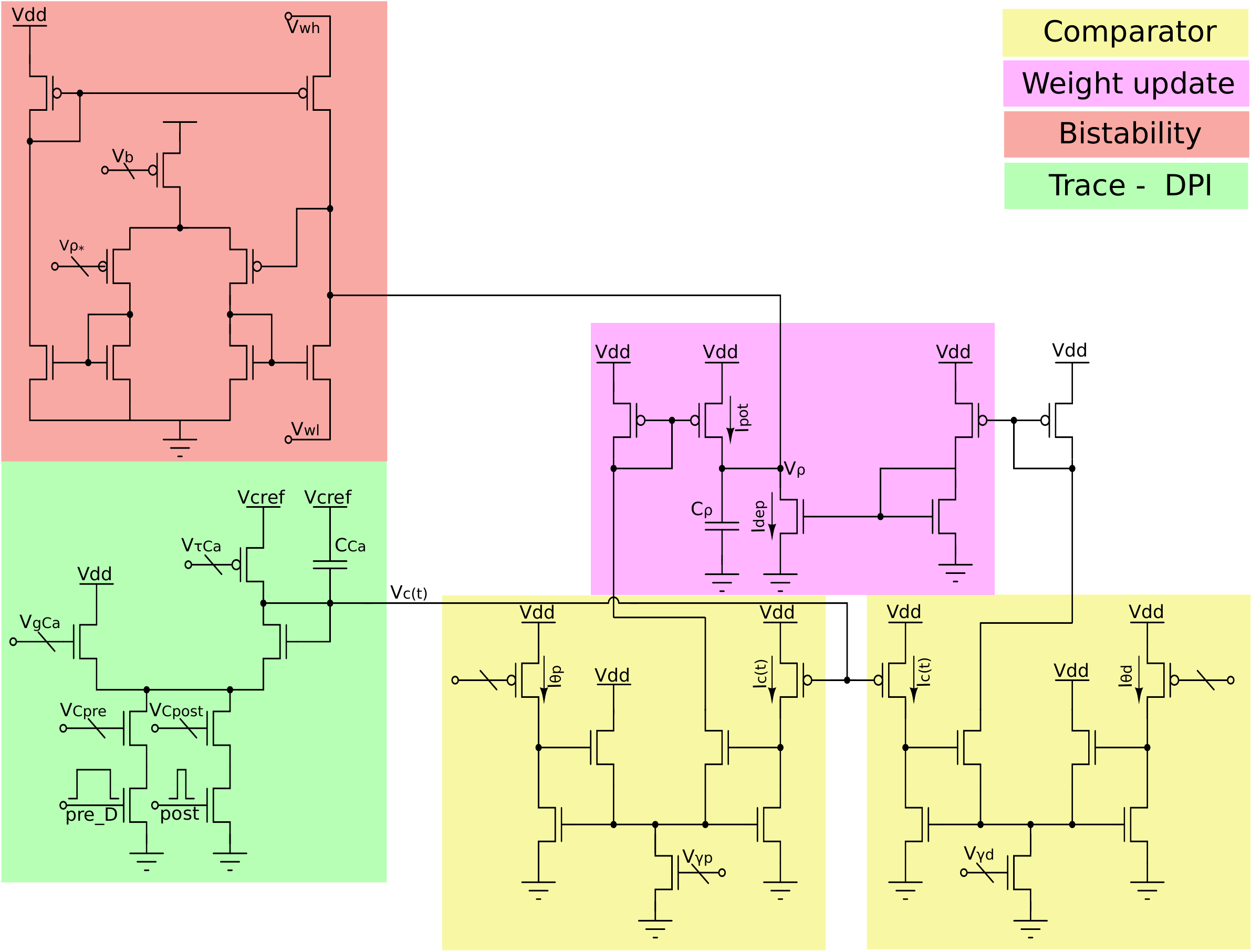}
        \centering
        \caption{\ac{CSTDP} circuit with highlighted \ac{CMOS} building blocks used: Eligibility traces with a \acs{DPI} (in green), Weight updates (in violet), Bistability (in red) and Comparators with \acs{WTA} (in yellow). Not shown is the circuit that implements the pre-synaptic spike extension. The voltage and current variables reflect the model equation. Adapted from:~\protect\citeasnoun{Maldonado_etal16}. 
        }
        \label{fig:cstdp}
    \end{figure}
\end{center}

The \ac{CSTDP} rule proposed by~\citeasnoun{Graupner_Brunel07} (see Eq.~\eqref{eq:cstdp}) attracted the attention of circuit designer thanks to its claim to closely replicate biological findings and explain synaptic plasticity in relation to both spike timing and rate. 
To implement the \ac{CSTDP} rule proposed by~\citeasnoun{Graupner_Brunel07} (see Eq.~\eqref{eq:cstdp}), \citeasnoun{Maldonado_etal16} made small adaptations to the original model and proposed the circuit shown in Fig.~\ref{fig:cstdp}.
Specifically, they proposed to convert the soft bounds of the efficacy update to hard bounds, resulting in the following model for the update of the synaptic efficacy:

\begin{equation}
    \begin{split}
        \tau \frac{d\rho}{dt} = -k_{bs}\rho(1 - &\rho)(\rho_{\star} - \rho) + \gamma_{p}\Theta[c(t) - \theta_p] - \gamma_{d}\Theta[c(t) - \theta_d] \\
        &\rho > 1 \rightarrow \rho = 1 \\
        &\rho < 0 \rightarrow \rho = 0
    \end{split}
\label{eq:ca_plasticity_simple}
\end{equation}

with $k_{bs}$ acting as a constant which scales the bistability dynamics and the hard-bounds implemented by the Heaviside function $\Theta$. 
The building blocks implemented in this work are shown in Fig.~\ref{fig:cstdp}. The trace block implements the local spike trace $c(t)$ represented by the voltage $V_{c}(t)$.
It consists of a \ac{DPI} with two input branches. On the arrival of either a post-synaptic spike ($post$) or the delayed pre-synaptic spike ($pre\_D$) the capacitor $C_{ca}$ is charged by a current defined by the gain of the \ac{DPI} ($V_{gCa}$) and $V_{Cpost}$ or $V_{Cpre}$, respectively. Charging the capacitor decreases the voltage $V_{c}(t)$. In the absence of input pulses, the capacitor discharges at a rate controlled by $V_{\tau Ca}$ towards its resting voltage $V_{cref}$.
The voltage $V_{c}(t)$ of the trace block sets the amplitude of the current $I_{c}(t)$ within the comparator blocks (see Fig.~\ref{fig:cstdp}). The current $I_{c}(t)$ is compared with the potentiation and depression thresholds defined by the currents $I_{\theta p}$ and $I_{\theta d}$, respectively. The \ac{WTA} functionality of the comparator circuits implements the Heavyside functionality of the comparison of the local spike trace $c(t)$ with the thresholds for potentiation ($\theta_p$) and depression ($\theta_d$) in the model (see Eq.~\eqref{eq:cstdp}). 

While the Calcium current $I_{c}(t)$ is greater than the potentiation threshold current $I_{\theta p}$, the synapse efficacy capacitor $C_{\rho}$ within the weight update block (see Fig.\ref{fig:cstdp}) is continuously charged by a current defined by the parameter $V_{\gamma p}$. Similarly, as long as $I_{c}(t)$ is greater than the depression threshold current $I_{\theta d}$, $C_{\rho}$ is constantly discharged with a current controlled by $V_{\gamma d}$. The voltage across the synapse capacitor $V_{\rho}$ resembles the efficacy $\rho$ of the synapse. 
To implement the bistability behavior of the synaptic efficacy, Maldonado et al. use an \ac{TA} in positive feedback configuration with a very small gain defined by $V_{b}$ (see Fig.~\ref{fig:cstdp}). As long as the synaptic efficacy voltage $V_{\rho}$ is above the bistability threshold $V_{\rho_{\star}}$ the positive feedback constantly charges the capacitor $C_{\rho}$ and drives $V_{\rho}$ towards the upper limit defined by $V_{wh}$. In the case that $V_{\rho}$ is below $V_{\rho_{\star}}$, the \ac{TA} discharges the capacitor and drives $V_{\rho}$ toward the lower limit defined by $V_{wl}$.


\subsection{\acf{RDSP}}
\begin{center}
    \begin{figure}[H]
        \includegraphics[width=0.7\textwidth,  angle =0]{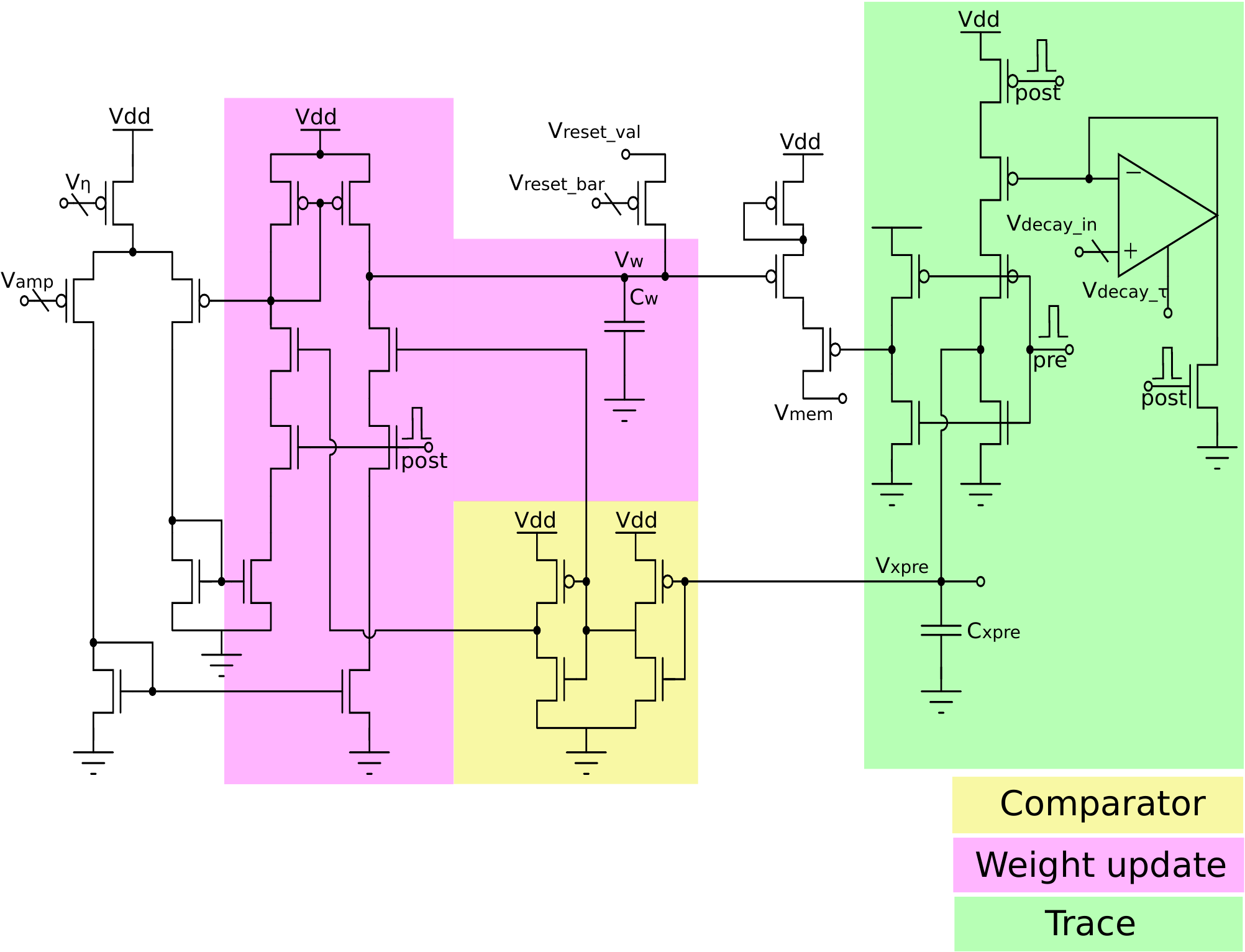}
        \centering
        \caption{\acs{RDSP} circuit with highlighted the \ac{CMOS} building blocks used: Eligibility traces (in green), weight updates (in violet) and comparators with differential pair (in yellow). Adapted from:~\protect\citeasnoun{Hafliger_etal97}.
       }
        \label{fig:rdsp}
    \end{figure}
\end{center}

The first \ac{CMOS} implementation of a spike-based learning rule done by~\citeasnoun{Hafliger_etal97} pre-dates the formalization of the \ac{RDSP} model, which happened almost 20 years later~\cite{Diehl_Cook15}. It is one of the most apparent cases of how building electronic circuits that mimic biological behavior leads to the discovery of useful mechanisms for solving real-world problems.

The algorithmic definition of their learning rule is based on a correlation signal, local to each synapse, which keeps track of the pre-synaptic spike activity. The correlation signal is refreshed at each pre-synaptic event and decays over time. When a post-signal arrives, depending on the value of the correlation, the weight is either increased or decreased, while the correlation signal is reset.
Similarly, the \ac{RDSP} rule relies on the pre-synaptic spike time information and is triggered when a post synaptic spike arrives. The direction of weight update depends on a target value $x_{tar}$, which determines the threshold between depression and potentiation.

The two main differences between the circuit by~\citeasnoun{Hafliger_etal97} (see Fig.~\ref{fig:rdsp}) and the \ac{RDSP} rule (see Eq.~\eqref{eq:rdsp}) is that the correlation signal in~\citeasnoun{Hafliger_etal97} is binary and is compared to a fixed threshold voltage (the switching threshold of the first inverter), which resembles a fixed $x_{tar}$.
In the~\citeasnoun{Hafliger_etal97} implementation, the voltage $V_{w}$ across the capacitor $C_{w}$ represents the synaptic weight and the voltage $V_{xpre}$ at the capacitor $C_{xpre}$ represents the correlation signal. At the arrival of a pre-synaptic input spike ($pre$), the voltage $V_{w}$ determines the amplitude of the current towards the soma ($V_{mem}$) of the post-synaptic neuron. At the same time, the capacitor $C_{xpre}$ is fully discharged and $V_{xpre}$ is low. In the absence of pre-synaptic and post-synaptic spikes ($pre$ and $post$ are low), $C_{xpre}$ is slowly charged towards $Vdd$ by the pMOS branch in the trace block (see Fig.~\ref{fig:rdsp}).

The voltage $V_{xpre}$ is constantly compared to the threshold voltage (resembling $x_{tar}$) of the first inverter it is connected to. At the arrival of a post-synaptic spike ($post$ is high) the weight capacitor $C_{w}$ is either charged (depressed) or discharged (potentiated) depending on the momentary level of $V_{xpre}$. If $V_{xpre}$ is above the inverter threshold voltage, the right branch of the weight update block (see Fig.~\ref{fig:rdsp}) is inactive, while the left branch is active and the pMOS-based current mirror charges the capacitor $C_{w}$. In the opposite case, where $V_{xpre}$ is below the inverter threshold voltage, the right branch is active while the output of the second inverter disables the left branch of the weight update block. This results in a discharge of the capacitor $C_{w}$ controlled by the nMOS-based current mirror. The amplitude for potentiation and depression is set by the two biases $V_{\eta}$ and $V_{amp}$. At the end of a post-synaptic spike the correlation signal $V_{xpre}$ is reset to $Vdd$.
A similar approach implementing a nearest-spike interaction scheme and a fixed $x_{tar}$ was implemented by~\citeasnoun{Ramakrishnan_etal11} exploiting the properties of floating gates.


\subsection{Other models implementations}
\label{sec:cmos_other}
To the best of our knowledge, there have been no dedicated \ac{CMOS}-based implementations of the other models presented in Sec.~\ref{sec:models}. Although the \ac{VSTDP} rule proposed by ~\citeasnoun{Clopath_etal10} and~\citeasnoun{Clopath_Gerstner10} shares similarities with the \ac{TSTDP} rule and can be related to the \ac{BCM} rule~\cite{Gjorgjieva_etal11}, its complexity for implementations comes from its multiple transient signals on different timescales. To this end, emerging novel technologies, such as memristors~\cite{Cantley_etal11,Li_etal13,Li_etal14,Ziegler_etal15,Diederich_etal18} and neuristors~\cite{Abraham_etal18} are capable of supporting promising solutions to implement different timescales in a compact and efficient manner. 
Similarly, implementations for the \ac{DPSS} rule~\cite{Urbanczik_Senn14} are difficult due to the increased complexity of the required multi-compartment neuron models. Recently, implementations based on hybrid memristor-\ac{CMOS} systems~\cite{Nair_etal17,Payvand_etal20} or using existing neuromorphic processors to exploit neuron structures to replicate the multi-compartment model~\cite{Cartiglia_etal20} have been proposed.
A detailed view on these implementations is beyond the scope of this review and the authors refer the readers to the original publications.

However, introducing \ac{CMOS} implemented models through the lens of functional building blocks allows us to quickly look for analogies and differences between the implemented and other models. Throughout this Section, we have highlighted the similarities and differences of each of the implemented models.
Focusing on functional building blocks also allows for a broader generalization to all the models that have not been implemented yet: using the basic building block we presented (e.g.\ Traces, Comparators, Weight updates, and Bistability) one could potentially construct all the learning models we have discussed in Sec.~\ref{sec:models}.


\section{Discussion and conclusion}
\subsection{Toward a unified synaptic plasticity framework}
In this survey, we highlighted the similarities and differences of representative synaptic plasticity models and provided examples of neuromorphic circuits \ac{CMOS} that can be used to implement their principles of computation. 
We highlighted how the principle of locality in learning and neural computation in general is fundamental and enables the development of fast, efficient and scalable neuromorphic processing systems. 
We highlighted how the different features of the plasticity models can be summarized in (1) synaptic weights properties, (2) plasticity update triggers and (3) local variables that can be exploited to modify the synaptic weight (see also Table~\ref{tab:models}). Although all local variables of these rules are similar in nature, the plasticity rules can can be subdivided in the following way:

\begin{itemize}
    \item Pre-synaptic spike trace: \ac{RDSP}.
    \item Pre- and post-synaptic spike traces: \ac{STDP}, \ac{TSTDP}, \ac{CSTDP}, \ac{SBCM}, \ac{BDSP}.
    \item Pre-synaptic spike trace + post-synaptic membrane voltage: \ac{VSTDP}, \ac{DPSS}, \ac{MPDP}, \ac{HMPDP}.
    \item Post-synaptic membrane voltage + post-synaptic spike trace: \ac{SDSP}, \ac{CMPDP}.
\end{itemize}

Many possibilities arise when exploring how the local variables used by these rules interact (e.g.\ comparison, addition, multiplication, etc.). This leads to a wide range of additional models that could be proposed and to a large number of biological experiments that could be carried out to verify the hypotheses and predictions made by the rules.

It is difficult to predict whether a unified rule of synaptic plasticity can be formulated, based on the observation that several plasticity mechanisms coexist in the brain~\cite{Abbott_Nelson00,Bi_Poo01}, and that different problems may require different plasticity mechanisms. 
Nevertheless, we provided here a single unified framework that allowed us to do a systematic comparison of the features of many representative models of synaptic plasticity presented in the literature, developed following experiment-driven bottom-up approaches and/or application-driven top-down approaches~\cite{Frenkel_etal21b}. While the bottom-up approach can help in explaining the plasticity mechanisms found in the brain, top-down guidance can help to find the right level of abstraction from biology to get the best performance for solving problems in the context of efficient and adaptive artificial systems. In line with the neuromorphic engineering perspective, this work bridges the gap between both approaches.


\subsection{Overcoming back-propagation limits for online learning}
\label{sec:gradient-learning}
Local synaptic plasticity in neuromorphic circuits offers a promising solution for online learning in embedded systems. However, due to the very local nature of this approach, there is no direct way of implementing global learning rules in multi-layer neural networks, such as the gradient-based back-propagation algorithm~\cite{LeCun_etal98,Schmidhuber_etal07}. 
This algorithm has been the work horse of \acp{ANN} training in deep learning over the last decade. Gradient-based learning has recently been applied for offline training of \acp{SNN}, where the \ac{BP} algorithm coupled with surrogate gradients is used to solve two critical problems: first, the temporal credit assignment problem which arises due to the temporal inter-dependencies of the \ac{SNN} activity. It is solved offline with \ac{BPTT} by unrolling the \ac{SNN} like standard \acp{RNN}~\cite{Neftci_etal19}. Second, the spatial credit assignment problem, where the credit or ``blame'' with respect to the objective function is assigned to each neuron across the layers.
However, \ac{BPTT} is not biologically plausible~\cite{Bengio_etal15,Lillicrap_etal20} and not practical for on-chip and online learning due to the non-local learning paradigm. 
On one hand, \ac{BPTT} is not local in time as it requires keeping all the network activities for the duration of the trial. 
On the other hand, \ac{BPTT} is not local in space as it requires information to be transferred across multiple layers. Indeed, synaptic weights can only be updated after complete forward propagation, loss evaluation, and back-propagation of error signals, which lead to the so-called ``locking effect''~\cite{Czarnecki_etal17}. 


Recently, intensive research in neuromorphic computing has been dedicated to bridge the gap between back-propagation and local synaptic plasticity rules by reducing the non-local information requirements, at a cost of accuracy in complex problems~\cite{Eshraghian_etal21}.
The temporal credit assignment can be handled by using eligibility traces~\cite{Zenke_Ganguli18,Bellec_etal20} that solve the distal reward problem by bridging the delay between the network output and the feedback signal that may arrive later in time~\cite{Izhikevich07}.
Similarly, inspired by recent progress in deep learning, several strategies have been explored to solve the spatial credit assignment problem using feedback alignment~\cite{Lillicrap_etal16}, direct feedback alignment~\cite{Nokland16}, random error \ac{BP}~\cite{Neftci_etal17} or by replacing the backward pass with an additional forward pass whose input is modulated with error information~\cite{Dellaferrera_Kreiman22}. 
However, these approaches only partially solve the problem~\cite{Eshraghian_etal21}, since they still suffer from the locking effect, which can nonetheless be tackled by replacing the global loss by a number of local loss functions~\cite{Mostafa_etal18,Neftci_etal19,Kraiser_etal20,Halvagal_Zenke22} or by using direct random target projection~\cite{Frenkel_etal21b}. 
Assigning credit locally, especially within recurrent \acp{SNN}, is still an open question and an active field of research~\cite{Christensen_etal21}. 

The local synaptic plasticity models and circuits presented in this survey do not require the presence of a teacher signal and contrast with supervised learning using labeled data which is neither biologically plausible~\cite{Halvagal_Zenke22} nor practical in most online scenarios~\cite{Muliukov_etal22}. Nevertheless, the main limit of spike-based local learning is the diminished performance on complex pattern recognition problems. Different approaches have been explored to bridge this gap, such as \ac{DPSS}~\cite{Urbanczik_Senn14,Sacramento_etal18} and \ac{BDSP}~\cite{Payeur_etal21} learning rules that use multi-compartment neurons and show promising performance in approximating back-propagation with local mechanisms, or using multi-modal association to improve the self-organizing system's performance~\cite{Gilra_Gerstner17,Khacef_etal20,Rathi_Roy21} as in contrast to labeled data, multiple sensory modalities (e.g.\ sight, sound, touch) are freely available in the real-world environment.


\subsection{Structural plasticity and network topology}
Exploring local synaptic plasticity rules gives valuable insights into how plasticity and learning evolves in the brain. However, in bringing the plasticity of single synapses to the function of entire networks, many more factors come into play. Functionality at a network level is determined by the interplay between the synaptic learning rules, the spatial location of the synapse, and the neural network topology. 

Furthermore, the network topology of the brain is itself plastic~\cite{Holtmaat_Svoboda09}. 
\citeasnoun{LeBe_Markram06} provided the first direct demonstration of induced rewiring (i.e.\ sprouting and pruning) of a functional circuit in the neocortex~\cite{Markram_etal11}, which requires hours of general stimulation.
Some studies suggest that glutamate release is a key determinant in synapse formation~\cite{Engert_Bonhoeffer99,Kwon_Sabatini11}, but additional investigations are needed to better understand the computational foundations of structural plasticity and how it is linked to the synaptic plasticity models we reviewed in this survey. Together, structural and synaptic plasticity are the local mechanisms that lead to the emergence of the global structure and function of the brain. Understanding, modeling, and implementing the interplay between these two forms of plasticity is a key challenge for the design of self-organizing systems that can get closer to the unique efficiency and adaptation capabilities of the brain.


\subsection{\ac{CMOS} neuromorphic circuits}
The computational primitives that are shared by the different plasticity models were grouped together in corresponding functional primitives and circuit blocks that can be combined to map multiple plasticity models into corresponding spike-based learning circuits.
Many of the models considered rely on exponentially decaying traces. By operating the \ac{CMOS} circuits in the sub-threshold regime, this exponential dependency is given by the physical substrate of transistors showing an exponential relationship between current and voltage~\cite{Mead90}.



The circuits presented make use of both analog computation (e.g.\ analog weight updates) and digital communication (e.g.\ pre- and post-synaptic spike events). This mixed-signal analog/digital approach aligns with the observations that biological neural systems can be considered as hybrid analog and digital processing systems~\cite{Sarpeshkar98}.
Due to the digital nature of spike transmission in these neuromorphic systems, plasticity circuits that require the use of pre-synaptic traces need extra overhead to generate this information directly at the post-synaptic side. 

The emergence of novel nanoscale memristive devices has high potential for allowing the implementation of such circuits at a low overhead cost, in terms of space and power~\cite{Demirag_etal21}.
In addition, these emerging memory technologies have the potential of allowing long-term storage of the synaptic weights in a non-volatile way, that would allow these neuromorphic systems to operate continuously, without having to upload the neural network parameters at boot time. This will be a significant advantage in large-scale systems, as Input/Output operations required to load network parameters can take a significant amount of power and time.
In addition, the properties of emerging memristive devices could be exploited to implement different features of the plasticity models proposed~\cite{Diederich_etal18}.

Overall, the number of proposed \ac{CMOS}-based analog or mixed-signal neuromorphic circuits over the past 25 years is relatively low, as this was mainly driven by fundamental academic research. With the increasing need for low-power neural processing systems at the edge, the increasing maturity of novel technologies, and the rising interest in brain-inspired neural networks and learning for data processing, we can expect an increasing number of new mixed signal analog/digital circuits implementing new plasticity rules also for commercial exploitation. In this respect, this review can provide valuable information for making informed modeling and circuit design decision in developing novel spike-based neuromorphic processing systems for online learning.


\ack

We would like to thank the BICS group for the fruitful discussions, with special thanks to Hugh Greatorex and Carver Mead for providing valuable feedback on the manuscript. We thank Wouter Serdijn for the fruitful discussions on circuit details. We would also like to acknowledge the financial support of the CogniGron research center and the Ubbo Emmius Funds (Univ. of Groningen), the European Union's H2020 research and innovation programme under the H2020 BeFerroSynaptic project (871737), the  Swiss National Science Foundation Sinergia project (CRSII5-18O316), and the ERC grant NeuroAgents (724295).


\section*{Data availability statement}
No new data were created or analyzed in this study.


\section*{ORCID IDs}

Lyes Khacef: https://orcid.org/0000-0002-4009-174X. \\
Philipp Klein: https://orcid.org/0000-0003-4266-2590. \\
Matteo Cartiglia: https://orcid.org/0000-0001-8936-6727. \\
Arianna Rubino: https://orcid.org/0000-0002-5036-1969. \\
Giacomo Indiveri: https://orcid.org/0000-0002-7109-1689. \\
Elisabetta Chicca: https://orcid.org/0000-0002-5518-8990.


\section*{References}

\bibliography{biblio/biblio.bib, biblio/bics.bib}

\end{document}